\documentclass[runningheads]{llncs}
\usepackage{graphicx}

\usepackage{tikz}
\usepackage{comment}
\usepackage{amsmath,amssymb} 
\usepackage{color}

\usepackage{bm}
\usepackage{float}
\usepackage{subcaption}
\usepackage{multirow}
\usepackage{cite}
\usepackage{enumitem}
\usepackage[font=small]{caption}
\usepackage{booktabs}
\usepackage[accsupp]{axessibility}  

\begin{document}
\pagestyle{headings}
\mainmatter
\def\ECCVSubNumber{5142}  

\title{DCCF: Deep Comprehensible Color Filter Learning Framework for High-Resolution Image Harmonization} 

\titlerunning{DCCF: Deep Comprehensible Color Filter Learning Framework}
%
\author{Ben Xue\inst{1}\thanks{Finish this work during an internship at Alibaba Group. Email: \email{xueben@pku.edu.cn}.}\and
Shenghui Ran\inst{2} \and
Quan Chen\inst{2}\thanks{Corresponding author. Email: \email{myctllmail@163.com}. }\and \\
Rongfei Jia\inst{2}\and
Binqiang Zhao\inst{2}\and
Xing Tang\inst{2}\\
}
\authorrunning{Xue et al.}
%
\institute{
Academy for Advanced Interdisciplinary Studies, Peking University, China \and
Alibaba Group, China}
\maketitle

\begin{abstract}

    Image color harmonization algorithm aims to automatically match the color distribution of foreground and background images captured in different conditions.
    Previous deep learning based models neglect two issues that are critical for practical applications, namely high resolution (HR) image processing and model comprehensibility.
    In this paper, we propose a novel Deep Comprehensible Color Filter (DCCF) learning framework for high-resolution image harmonization.
    Specifically, DCCF first downsamples the original input image to its low-resolution (LR) counter-part, then learns four human comprehensible neural filters (i.e. hue, saturation, value and attentive rendering filters) in an end-to-end manner, finally applies these filters to the original input image to get the harmonized result.
    Benefiting from the comprehensible neural filters, we could provide a simple yet efficient handler for users to cooperate with deep model to get the desired results with very little effort when necessary.
    Extensive experiments demonstrate the effectiveness of DCCF learning framework and it outperforms state-of-the-art post-processing method on iHarmony4 dataset on images' full-resolutions by $7.63\%$ and $1.69\%$ relative improvements on MSE and PSNR, respectively. Our code is available at \url{https://github.com/rockeyben/DCCF}.

    \end{abstract}
    
    \section{Introduction}

\begin{figure}
        \centering
        \includegraphics[width=0.98\linewidth]{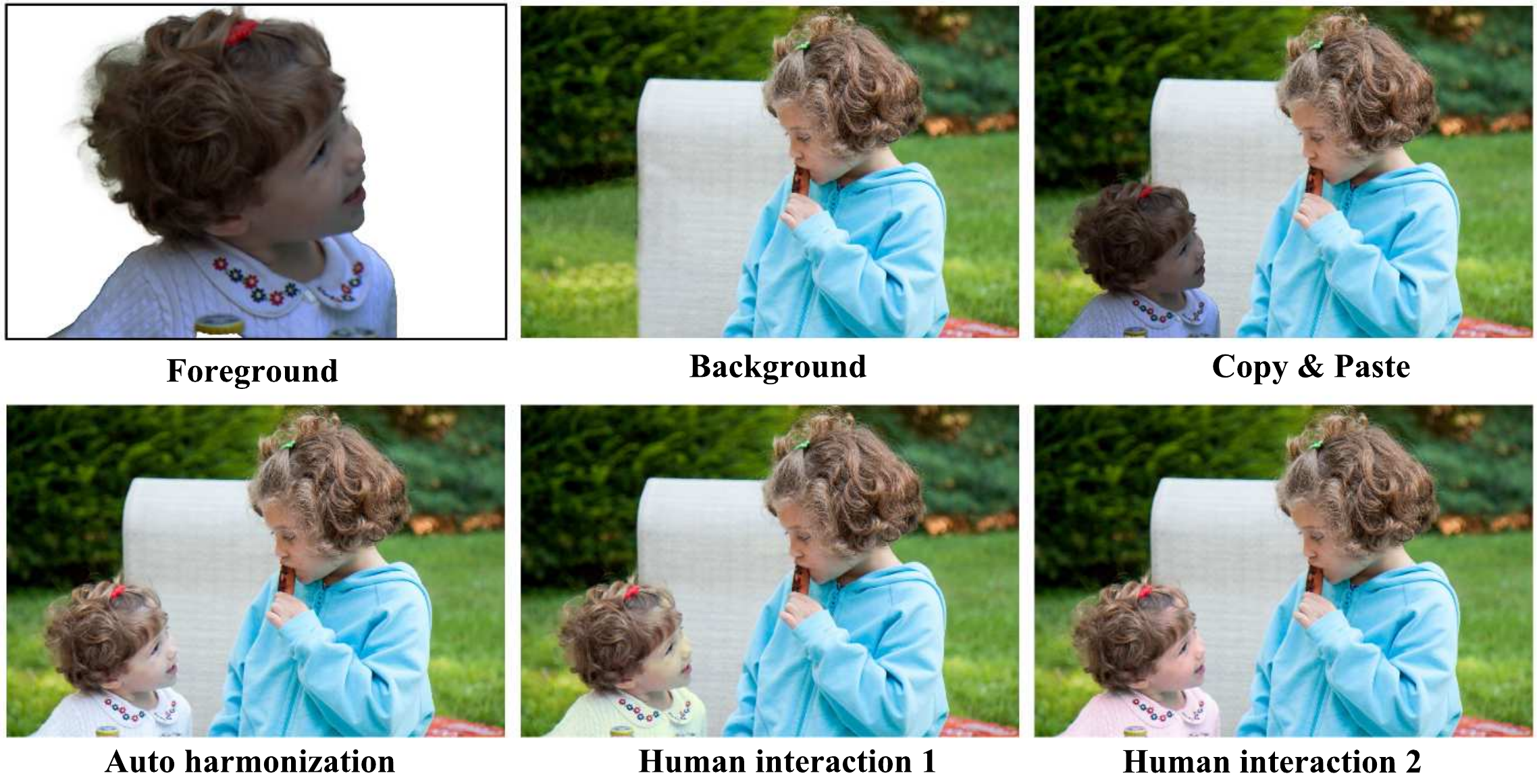}\\
        \caption{
            \small Illustration of color harmonization}
        \label{fig:color_harmonization_demo}
        \vspace{-0.8cm}
    \end{figure}

    Image composition, which aims at generating a realistic image with the given foreground and background, is one of the most widely used technology in photo editing.
    However, since the foreground and background may be captured in different conditions, simple cutting and pasting operations could not make them compatible in color space, as show in Fig.~\ref{fig:color_harmonization_demo}.
    Therefore, photo editors spend a lot of time in manual tuning the color distribution when they accomplish the real-world composition task.
    
    In the past decades, a large amount of automatic color harmonization algorithms have been proposed.
    Traditional methods~\cite{cohen2006color, pitie2005n, pitie2007linear, reinhard2001color, perez2003poisson, jia2006drag, tao2010error, sunkavalli2010multi} tend to extract low-level handcrafted features to make the color statistics of the foreground to match the background, which may have poor performance when the content of foreground and background are vastly different.
    Since Tsai et al.~\cite{tsai2017deep} propose a data-driven deep learning framework for color harmonization, the research community has made a large progress rapidly over a short period of time.
    Deep learning based methods have become the main stream.
    However, we argue that previous deep learning based color harmonization methods~\cite{tsai2017deep, cun2020improving, sofiiuk2021foreground, Ling_2021_CVPR, cong2020dovenet, cong2021bargainnet, guo2021intrinsic} have neglected two problems which are critical for practical applications.
    
    First, high-resolution (HR) images are rarely taken into account in previous works when deep color harmonization models are designed and evaluated.
    Previous deep models in color harmonization follow the evaluation system proposed by Tsai et al.~\cite{tsai2017deep}, which resizes the original images to 256$\times$256 or 512$\times$512 resolution and calculate objective metrics (i.e. MSE and PSNR) in this low-resolution to evaluate the performance of models, instead of the original image resolution.
    The principal reason is that these methods simply employ UNet-style~\cite{Unet} networks to directly predict pixel level RGB values, which are memory and computational costly, and even modern GPUs could not burden for HR images. 
    However, color harmonization needs to be frequently applied to HR images in real-world applications whose resolution is 3000$\times$3000 or even higher.
    Therefore, previous deep models which perform well on low-resolutions may have poor performance when be applied to real-world HR images.

    Second, model comprehensibility and manual control mechanism are rarely considered in previous works.
    Imagine the scenario that the harmonization result of the network is flawed, and the photo editor wants to make some modifications based on the network's prediction to avoid tuning from scratch, such as hue adjustment in Fig.~\ref{fig:color_harmonization_demo}.
    Thus it is essential to provide human understandable cooperation mode with the deep models for a friendly color harmonization system.
    However, previous methods utilize variant networks following the common image-to-image translation framework~\cite{isola2017image} that directly predicts the harmonization result.
    It is nearly impossible to provide comprehensible tools for humans to interact with these deep models, because of the prediction processes are "black-box" and inscrutable for photo editors.
    
    Inspired by the idea of learning desired image transformations that could reduce computing and memory burdens by a large margin for image enhancement~\cite{gharbi2017deep}, in this paper, 
    we propose a novel Deep Comprehensible Color Filter (DCCF) learning framework for high-resolution image harmonization.
    Specifically, we first downsample the input to the low-resolution (such as 256$\times$256) counter-part, then learn four comprehensible neural filters (i.e. hue, saturation, value and attentive rendering filters) in a novel end-to-end manner with the supervisions constructed from both RGB and HSV color spaces, finally apply these filters to the original input image to get the harmonized result.
    Compared with previous deep learning based color harmonization methods that may fail for high-resolution images, our neural filter learning framework is insensitive to image resolution and could perform well on dataset whose resolution range from 480p to 4K.
    Besides, benefiting from the mechanism that parameters in the filters (especially hue, saturation and value filters) are forced to learn decoupled meaningful chromatics functions, it makes it possible to
    provide comprehensible tools for humans to interact with these deep models in the traditional chromatics way they familiar with.
    It is worth noting that learning comprehensible neural filters is not easy.
    Our experiments show that learning weights directly from supervisions of hue, saturation and value channels could cause poor performance.
    To handle this, we construct three novel supervision maps that approximate the effects of HSV color space while making the deep model converge well.
    
    We train and evaluate our approach in the open source iHarmony4 dataset ~\cite{cong2020dovenet} on the original image resolutions, which range from 480p (HCOCO) to 4K (HAdobe5k).
    Since previous deep learning based color harmonization models could perform poorly when they are directly applied to HR images, we compare to them with variant post-processing methods.
    Extensive experiments demonstrate that our approach can make the prediction process comprehensible and outperform these methods as well.
    We also provide a simple handler that humans could cooperate with the learned deep model to make some desired modifications based on the network's prediction capacity to avoid tuning from scratch. 
    
    In a nutshell, our contributions are three-folds.
    \begin{itemize}[noitemsep,topsep=0pt]
        \item
        We propose an effective end-to-end deep neural filter learning framework that is insensitive to image resolution, which makes deep learning based color harmonization practical for real-world high-resolution images.
        
        \item
        To the best of our knowledge,  we are the first to design four types of novel neural filter (i.e. hue, saturation, value and attentive rendering filters) learning functions and learning strategies that make the prediction process and result comprehensible for human in image harmonization task .
        Meanwhile, we provide a simple yet efficient handler for users to cooperate with deep model to get the desired results with very little effort when necessary.
        
        \item Our approach achieves state-of-the-art performance on the color harmonization benchmark for high-resolution images and outperforms state-of-the-art post-processing method by $7.63\%$ and $1.69\%$ relative improvements on MSE and PSNR, respectively.
    \end{itemize}

    
    
    \section{Related Work}
    
    \textbf{Image harmonization}.
    In this subsection, we focus on the discussion of deep learning based methods.
    These methods regard color harmonization as a black box image-to-image translation task.~\cite{tsai2017deep} apply the well-known encoder-decoder  U-net structure with skip-connection and train the network with multi-task learning, simultaneously predicting pixel value and semantic segmentation.
     ~\cite{sofiiuk2021foreground} insert pretrained semantic segmentation branch into encoder backbone and introduce a learnable alpha-blending mask to borrow useful information from input image. They both use semantic features in networks.~\cite{cong2020dovenet, cong2021bargainnet} tried to make composite image harmonious via domain transfer.
    ~\cite{cun2020improving, hao2020image} both used attention mechanism in networks.
    ~\cite{chen2019toward} propose a generative adversarial network (GAN) architecture for automatic image compositing, which considers geometric, color, and boundary consistency at the same time.
    ~\cite{guo2021intrinsic} seek to solve image harmonization via separable harmonization of reflectance and illumination, where reflectance is harmonized through material-consistency penalty and illumination is harmonized by learning and transferring light from background to foreground.
    Note that recently some image harmonization works start to focus on high-resolution images.
    ~\cite{jiang2021ssh} use self-supervised learning strategy to train network with small local patches of high resolution images, but during inference it still follow the two stage post-processing strategy.
    ~\cite{hu2018exposure,shi2021learning} learn global parameters to adjust image attributes such as lightness and saturation. 
    ~\cite{Zero-DCE} learns pixel-wise curves to perform low-light image enhancement.
    
    \noindent\textbf{Smart upsampling}. Processing high resolution image becomes difficult due to huge computational burden of deep-learning networks and limited GPU memory.
    A common approach to accelerate high resolution processing is to first downsample the image, apply time-consuming operator at low resolution and upsample back. To preserve edge gradients, guided filter upsampling~\cite{hekaiming2013GF} uses original high resolution input as guidance map. 
    ~\cite{gharbi2015transform} fit transformation recipe from compressed input and output, then apply the recipe to high quality input. Bilateral guided upsampling~\cite{chen2016bilateral} approximates the operator with grids of local affine transformations and apply them on high resolution input, thus control the operator complexity. 
    ~\cite{gharbi2017deep} predict the local affine model with fully convolution networks, which is trained by end-to-end learning and obtain multi-scale semantic information. 
    ~\cite{wu2018fast} propose a guided filter layer, using point-wise convolution to approximate median filter, thus can be plugged into networks and optimized jointly. 
    ~\cite{kim2019deformable} introduce extra networks to learn deformable offsets for each pixel, thus the interpolation neighbour is predicted online during upsampling. 
    ~\cite{CDTNet,zeng2020learning} learn 3D lookup tables (LUT) to obtain high resolution results, but the learned transformation still lacks interpretable meanings.

    
    \begin{figure*}[h]
    \centering
    \includegraphics[width=\textwidth]{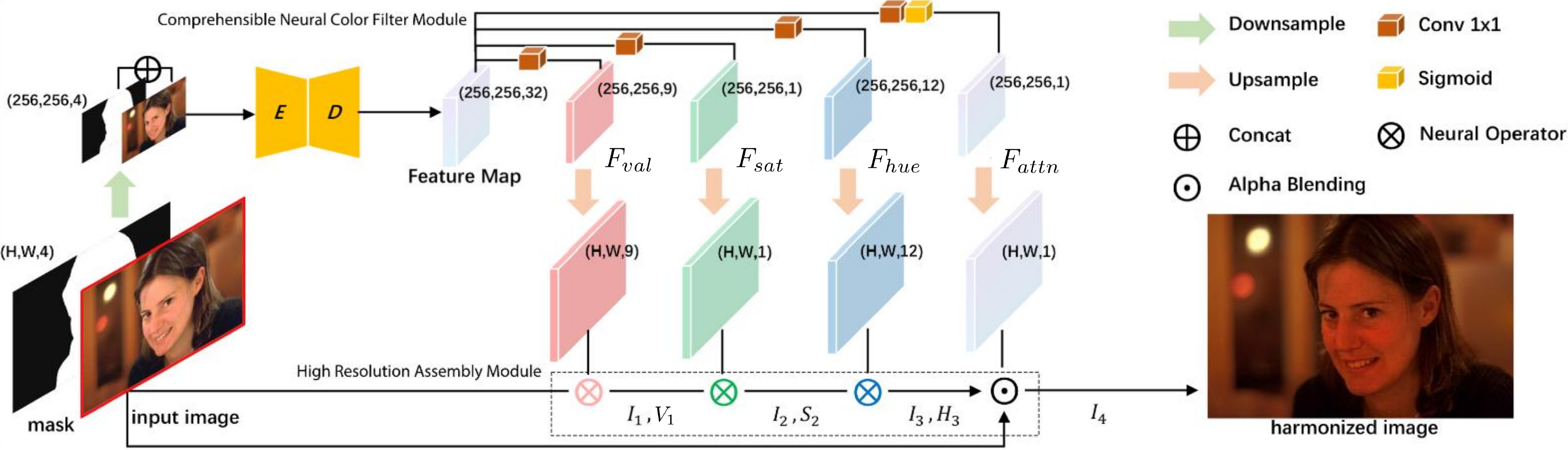}
    \caption{
    An overview of our proposed color harmonization framework. It consists of two primary parts: \textit{comprehensible neural color filter module} and \textit{high resolution assembly module}.
    Given an input image and corresponding foreground mask, a low-resolution feature extraction backbone first downsamples them to a low-resolution version, such as 256$\times$256, and employs an encoder-decoder network to extract foreground aware high-level semantic features.
    \textit{Comprehensible neural color filter module} then learns value filter, saturation filter, hue filter and attentive rendering filter simultaneously based on the features extracted from the backbone.
    Each filter learns parameters of transformation function in per pixel manner.
    \textit{High resolution assembly module} finally extracts and upsamples the specific channel of each DCCF's output to assemble the final result.
    In short, input image $I$ is unharmonious, $I_1$ is $V$-harmonized, $I_2$ is $V,S$-harmonized, $I_3$ is $V,S,H$-harmonized, $I_4$ is the refinement of $I_3$ by an attention module.
    }
    \label{fig:pipeline}
    \end{figure*}

    \section{Methodology}
    
    \subsection{Framework Overview}

    The neural filter learning framework for high-resolution image color harmonization is illustrated in Fig.~\ref{fig:pipeline}.
    It consists of two primary parts: \textit{comprehensible neural color filter module} and \textit{high resolution assembly module}.
    
    Firstly, given an original input image ($H$$\times$$W$$\times$3) and corresponding foreground mask ($H$$\times$$W$$\times$1), low-resolution feature extraction backbone downsamples them to the low-resolution counterparts (256$\times$256), then concatenates them as input (256$\times$256$\times$4) to extract foreground aware high-level semantic representations (256$\times$256$\times$32).
    The choice of backbone structure is flexible and iDIH-HRNet architecture~\cite{sofiiuk2021foreground} is used in this paper.
    
    Subsequently, the \textit{comprehensible neural color filter module} generates a series of deep comprehensible color filters (\textbf{DCCFs}) with the shape of (256$\times$256$\times$$D$), where each pixel has $D$ learnable parameters $\textbf{\textit{q}}=[q_1, q_2, ..., q_D]$ to construct a transformation function $f(I;\textbf{\textit{q}})$ which can be operated on input image $I$. 
    The gathering of each pixel's functions $f$ builds up a filter map $F$.
    The design of DCCFs and their cooperating mechanism will be detailed in Section~\ref{subsection:DCCF}.
    
    Finally, the \textit{high resolution assembly module} upsamples these filter maps to their full-resolution ($H$$\times$$W$) counterparts in order to be applied on the resolution of original input image. 
    Meanwhile, since each DCCF only changes a specific aspect of image, an assembly strategy is thus required to ensure there is no conflicts between each filter's operating procedure.
    The details will be discussed in Section~\ref{subsection:assembly}.

    
    The entire network is trained in an end-to-end manner and benefited from the supervision of the full-resolution images.
    Moreover, we observe that traditional losses in RGB color space is not sufficient for achieving state-of-the-art quality.
    We therefore propose auxiliary losses in Section~\ref{subsection:loss} for each DCCF's output to ensure that they are functioning as expected.

    \subsection{Comprehensible Neural Color Filter Module}
    \label{subsection:DCCF}
    
    The comprehensible neural color filter module plays a core rule in our proposed high-resolution image color harmonization framework.
    We take inspiration from the famous HSV color model which is widely used in photo editing community. 
    Compared with RGB color space, HSV is much more intuitive and easier for humans to interact with computers for color tuning.
    
    Our module consists of four neural filters, that is, \textit{value filter}, \textit{saturation filter}, \textit{hue filter} and \textit{attentive rendering filter} illustrated as $F_{val}$, $F_{sat}$, $F_{hue}$ and $F_{attn}$ respectively in Fig.~\ref{fig:pipeline}.
    Each filter is generated by a 1$\times$1 convolutional layer (expect for the attentive rendering filter has extra sigmoid layer for nomorlization) that builded on the low-resolution feature extraction backbone.

    \subsubsection{Value Filter} 
    

    \begin{figure}[!thb]
        \centering
        \vspace{-30pt}
        \includegraphics[width=\linewidth]{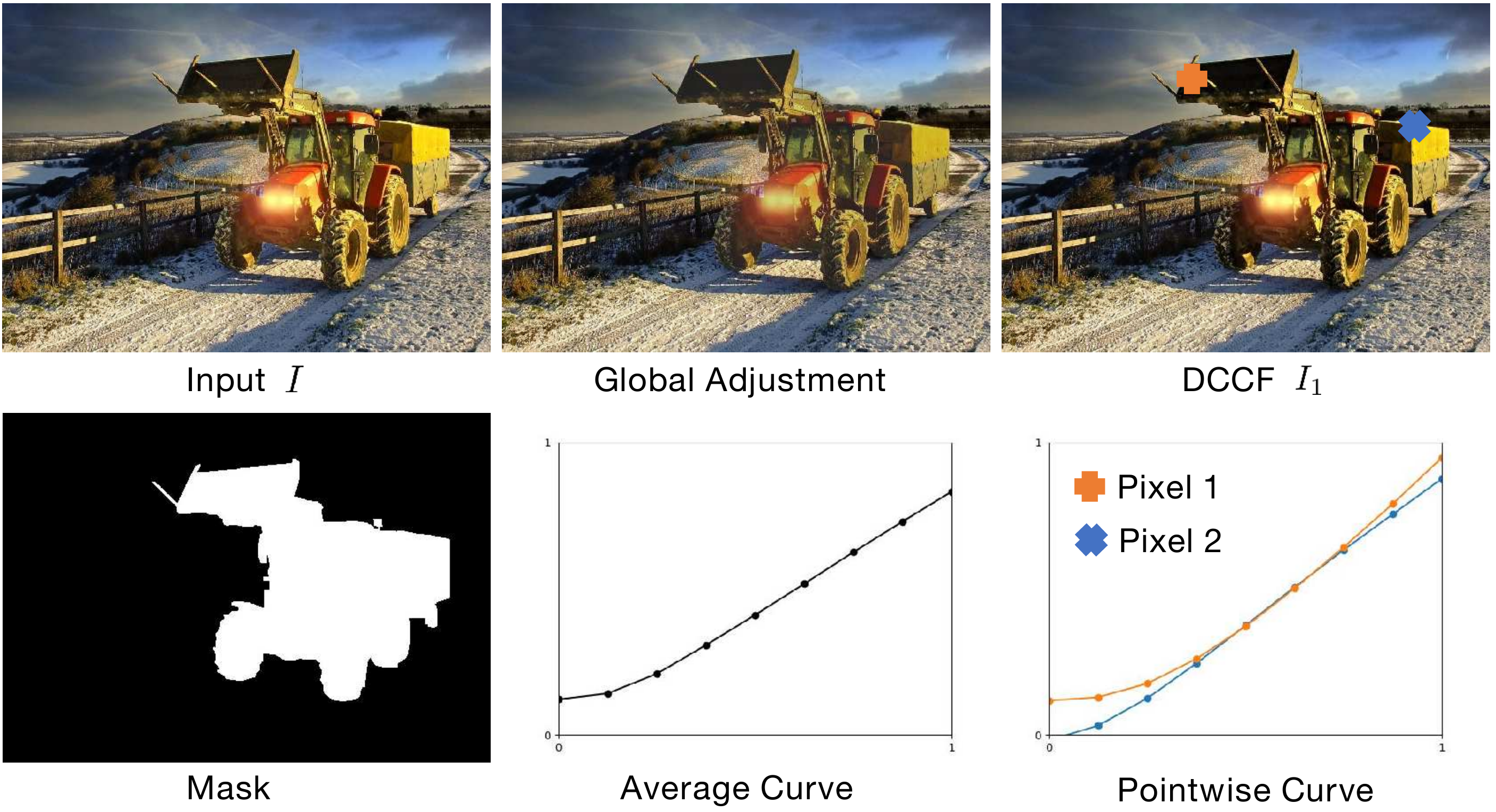}\\
        \caption{
            \small Illustration of the pixel-level value adjustment function/curve. $I_1=F_{val}(I)$ illustrated in Fig.~\ref{fig:pipeline} can be regarded as the result whose value is well tuned. Zoom for better view.
            }
        \vspace{-15pt}
        \label{fig:luminosity_curve}
    \end{figure}

    The customized pointwise nonlinear value transformation function $f_{val}$ is defined as:
     
    \begin{equation} \label{eq:f_lum}
        f_{val}(x;\phi,V_{min}) = V_{min} + \sum_{i=1}^{m} \phi_i * \max(x - \frac{i-1}{m}, 0)
    \end{equation}
    
    Where $x$ indicates the $V$ channel of input image in HSV color space,  $V_{min}$ and $\phi_i$ are learnable parameters and $m$ is a hyper-parameter which we set as 8 in this paper.
    It could be considered as an arbitrary nonlinear curve which is approximated by a stack of parameterized ReLUs.
    $V_{min}$ controls the lower-bound of value range, $m$ and $\phi_i$ control the nonlinearity of the curve.
    Parameters $V_{min}$ and $\phi_i (i=1,..,8)$ are stored for each pixel in channel direction of value filter $F_{val}$.
    
    
    We argue that different local regions should have different adjustment curves for better harmonization quality.
    As illustrated in Fig.~\ref{fig:luminosity_curve}, two marked points have large gap in original value distribution (the left is darker, the right is brighter), our DCCF $F_{val}$ successfully allocates proper curves for these two regions, while the global adjustment degrades the overall aesthetic.
    

    \subsubsection{Saturation Filter} 
    
    \begin{figure}[!thb]
        \centering
        \vspace{-30pt}
        \includegraphics[width=\linewidth]{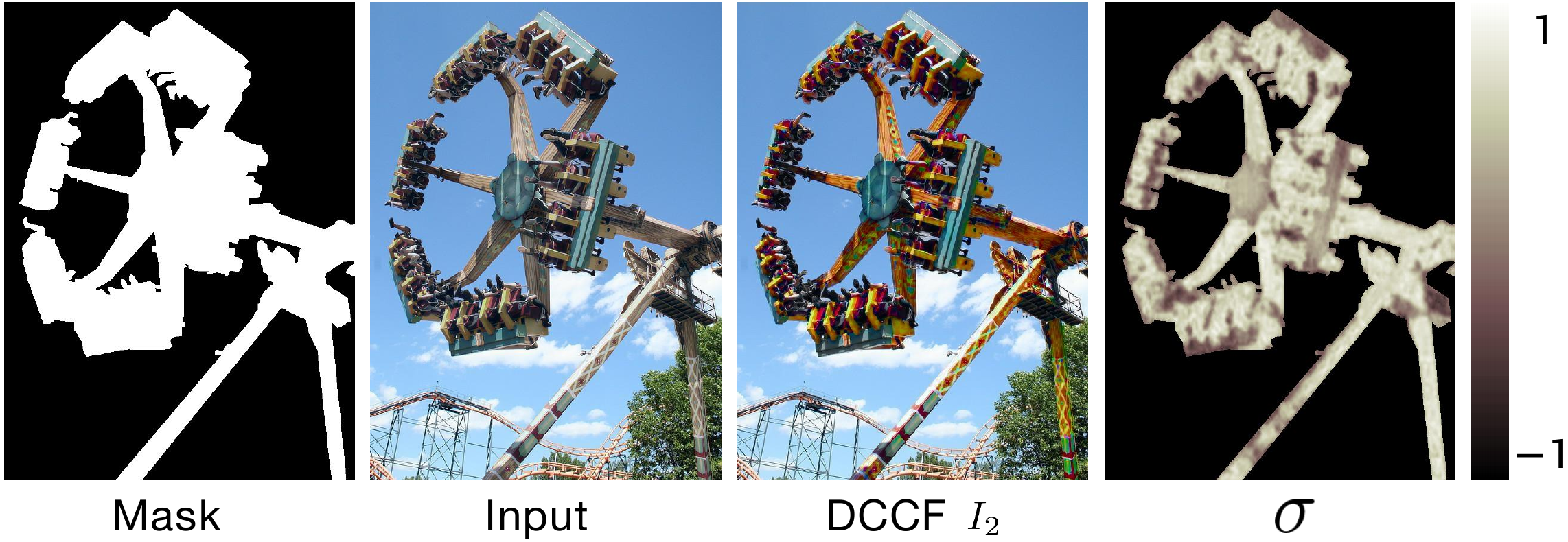}\\
        \caption{
            \small Illustration of the pixel-level saturation adjustment. 
            $I_2=F_{sat}(I_1)$ illustrated in Fig.~\ref{fig:pipeline} is the intermediate result. The change of saturation is consistent with predicted $\sigma$ distributions. Zoom for better view.
            }
        \vspace{-15pt}
        \label{fig:saturation_map}
    \end{figure}

    
    We use a single parameter $\sigma \in [-1,1]$ to control saturation for each pixel.
    The customized non-linear saturation transformation function $f_{sat}$ for each pixel is defined as:
    
    \begin{equation}
    \label{eq:sat}
    \begin{aligned}
            &f_{sat}(x;\sigma) = x + (x - C_{med}) * clip(\sigma) \\
    \end{aligned}
    \end{equation}
      
    Where $x$ indicates the $R$, $G$ or $B$ values in each pixel, $C_{max}=max(R,G,B)$, $C_{min}=min(R, G, B)$, $C_{med} = (C_{min}+C_{max})/2$,
    $\sigma$ is our learned parameter and $clip(\sigma)$ is a monotonous function to avoid saturation overflow.
    
    
    If $\sigma \rightarrow 1$, the values below median will be suppressed, while the value above median will be enhanced, as a result the saturation is increased and vice versa when $\sigma \rightarrow -1$.
    We visualize the effectiveness of $\sigma$ in Fig.~\ref{fig:saturation_map}. 
    DCCF allocated positive $\sigma$ for most of the pixels in this de-saturated input image and obtained an enhanced result.
    
    
    \subsubsection{Hue Filter}
    
    We define an affine color transformation function $f_{col}$ for each pixel in RGB color space as:
    
    %
    
    \begin{equation}
    \begin{aligned}
    f_{col}(x; \bm{\Delta})&=\bm{R}x + t \\
    &=
    \begin{bmatrix}
    \delta_{11} & \delta_{12} & \delta_{13} \\
    \delta_{21} & \delta_{22} & \delta_{23}\\
    \delta_{31} & \delta_{32} & \delta_{33}
    \end{bmatrix}
    \begin{bmatrix}
    x_R\\
    x_G\\
    x_B
    \end{bmatrix}
    +
    \begin{bmatrix}
    \delta_{14}\\
    \delta_{24}\\
    \delta_{34}
    \end{bmatrix}
    \end{aligned}
    \label{eq:hue_filter}
    \end{equation}
    
    Where $x$ indicates the RGB values for one pixel in image, and $\bm{\Delta}$ is a learnable 3x4 affine transformation matrix that contains a rotation matrix $\bm{R}$ and a translation vector $t$.
    
    We suppose that one could find a suitable rotation matrix $\bm{R}$ in RGB color space that is equivalent to a corresponding radian moving $\bm{r}$ on the hue ring in HSV color space~\cite{haeberli1993matrix}, which is futher discussed in supplementary.
    Based on this assumption, it is equivalent to learn an affine color transformation function $f_{col}(x; \bm{\Delta})$ in RGB color space, which contains a rotation function $\bm{R}$ that could be the parameters for the corresponding hue rotation function $f_{hue}(h; \bm{R})$ in HSV color space.
    We suggest readers refer to~\cite{haeberli1993matrix} for technical details. Note that~~\cite{haeberli1993matrix} needs extra linearization between sRGB and RGB space, which is mainly a gamma correction thus compatible with our learnable curve function $f_{val}$.
    
    
    \subsubsection{Attentive Rendering Filter} 
    We employ simple yet effective attentive rendering filter $F_{attn}$ which is similar to the attention mask in~\cite{sofiiuk2021foreground} to further improve the harmonization result after hue filter.
    
    
    For inference, we adopt the previous filters' harmonization result $I_3$ and input $I$ to perform alpha blending as illustrated in Fig.~\ref{fig:pipeline}
    \begin{equation}
        I_4 = I * \alpha + W_{ref} * I_3 * (1- \alpha)
    \end{equation}
    
    Where $\alpha$ is the per-pixel parameter on $F_{attn}$ ranging in $[0, 1]$ to smartly borrow information from input image, $W_{ref}$ is an extra affine matrix to refine the appearance of $I_3$.

    \subsection{High Resolution Assembly Module} \label{subsection:assembly}
    
    The biggest reduction of computation comes from the design that each DCCF is generated at low-resolution branch.
    We then perform upsampling on DCCF's filter map to match the resolution of original input image.
    The effectiveness of this action is guaranteed by the common assumption that neighbourhood regions require similar tuning filters. 
    
\begin{figure}[t]
        \centering
        \includegraphics[width=0.8\textwidth]{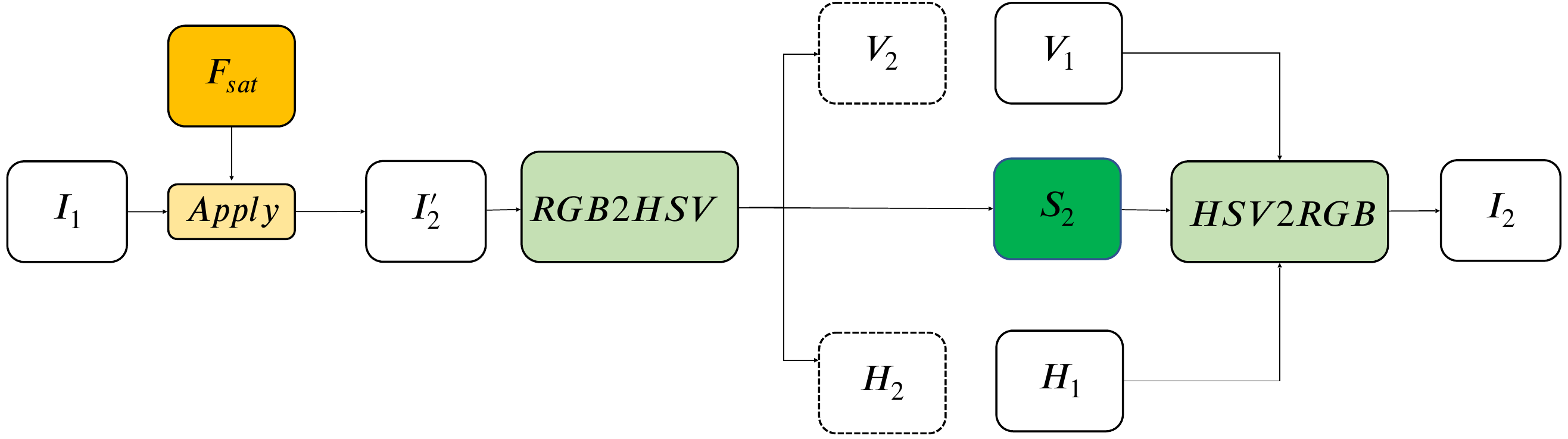}
        \caption{
        Illustration of assembly module details. We take the procedure of saturation filter $F_{sat}$ as example. The engaged channel (i.e. $S_2$) are colored for visualization.
        }
        \vspace{-15pt}
    \label{fig:pipeline_add}
    \end{figure}

    Afterwards, we propose a split-and-concat strategy to assemble the applying result of each filter. 
    Specifically, as shown in Fig.~\ref{fig:pipeline}, we utilize value filter $F_{val}$, saturation filter $F_{sat}$ and hue filter $H_{hue}$ to extract harmonized value channel $V_1$, saturation channel $S_2$ and hue channel $H_3$ respectively, then assemble $V_1$, $S_2$ and $H_3$ as harmonized image $I_3$, finally use attentive rendering filter to get the final harmonized image $I_4$.
    We illustrate the implementation details of saturation assembling as example in Fig.~\ref{fig:pipeline_add}.
    
    \subsection{Training Loss} \label{subsection:loss}
    
    In the following description, we will use the superscript $l$ for low-resolution and $h$ for high-resolution.
    
    
    \subsubsection{High Resolution Supervision}
    
    Since the area of foreground region varies a lot among training examples, we adopt foreground-normalized MSE loss~\cite{sofiiuk2021foreground} between ground-truth $I_{gt}$ and intermediate result $I_3$, final predicted result $I_4$.
    This loss uses the area of foreground mask as a normalization factor to stablize the gradient on foreground object.
    Differently, our loss can be calculated on both low-resolution and high-resolution streams, namely $\mathcal{L}_{rgb}^l$ and $\mathcal{L}_{rgb}^h$. 
    
    
    
    \subsubsection{Auxiliary HSV Loss} \label{section:smooth}
    


\begin{figure}[!thb]
        \centering
    \begin{minipage}[t]{0.35\textwidth}
        \centering
        \begin{subfigure}[t]{0.47\linewidth}
            \centering
            \includegraphics[width=1\linewidth]{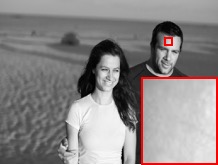}
            \caption{Standard V}\label{fig:std-l}
        \end{subfigure}
        \begin{subfigure}[t]{0.47\linewidth}
            \centering
            \includegraphics[width=1\linewidth]{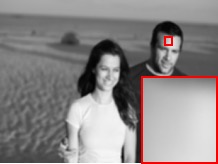}
            \caption{Smooth V}\label{fig:smo-l}		
        \end{subfigure}

        \begin{subfigure}[t]{0.47\linewidth}
            \centering
            \includegraphics[width=1\linewidth]{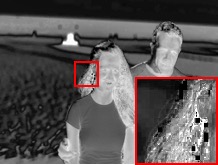}
            \caption{Standard S}\label{fig:std-s}
        \end{subfigure}
        \begin{subfigure}[t]{0.47\linewidth}
            \centering
            \includegraphics[width=1\linewidth]{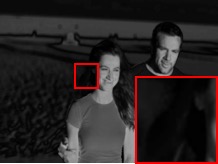}
            \caption{Smooth S}\label{fig:smo-s}
        \end{subfigure}

        \begin{subfigure}[t]{0.47\linewidth}
            \centering
            \includegraphics[width=1\linewidth]{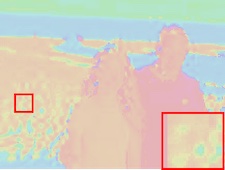}
            \caption{Standard H}\label{fig:std-h}
        \end{subfigure}
        \begin{subfigure}[t]{0.47\linewidth}
            \centering
            \includegraphics[width=1\linewidth]{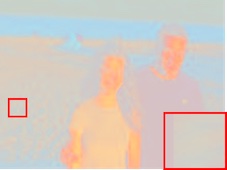}
            \caption{Smooth H}\label{fig:smo-h}
        \end{subfigure}
    \end{minipage}
    \begin{minipage}[t]{0.57\textwidth}
        \centering
        \vspace{-45pt}
        \begin{subfigure}[t] {1\linewidth}
            \centering
            \includegraphics[width=1\textwidth]{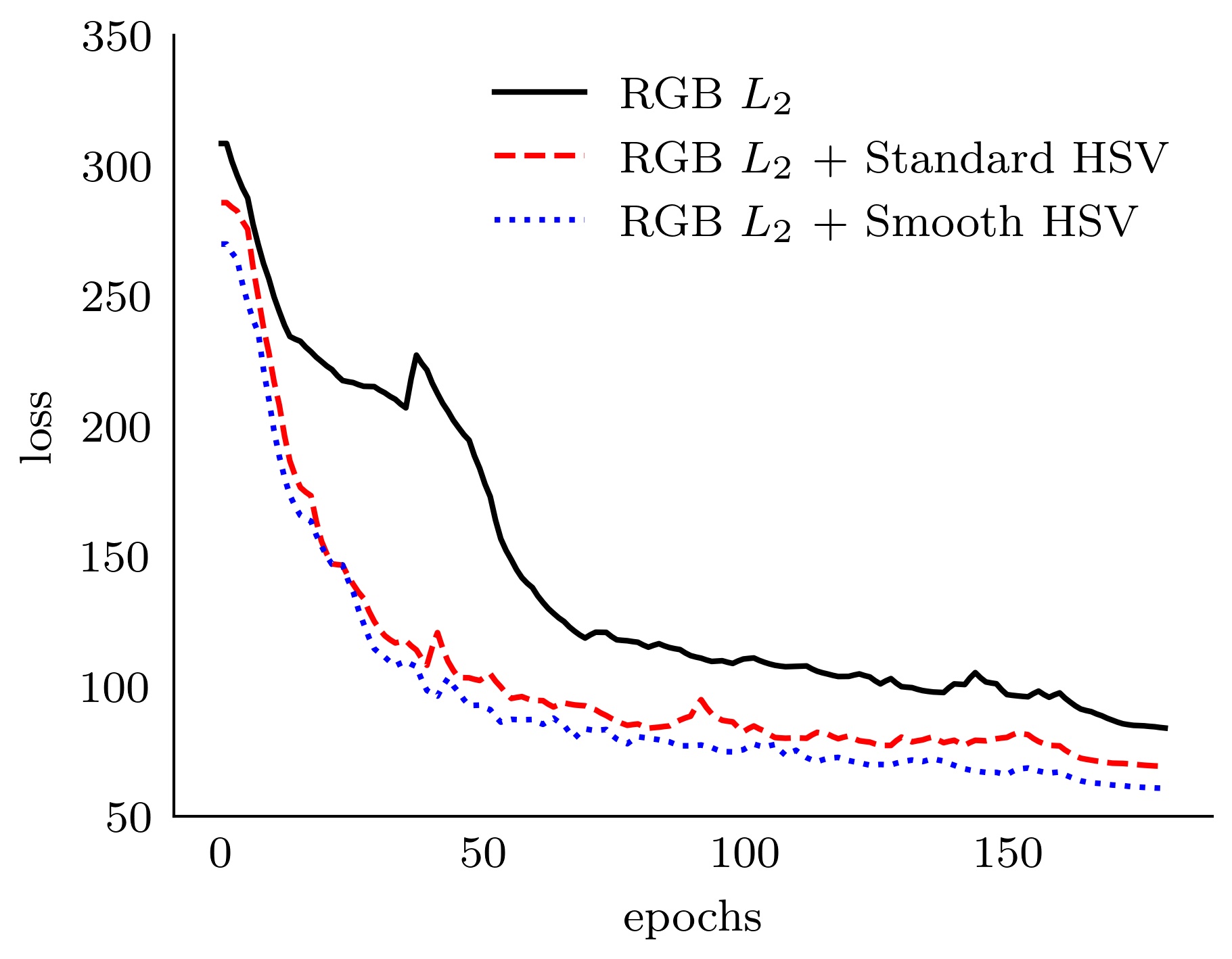}
            \caption{
                Ablations on loss function.
            Testing errors on iHarmony4 with different loss contributions.
            } 
            \label{fig:loss_curve}
        \end{subfigure}
    \end{minipage}
    
        \caption{
    Visualzation of standard HSV and our ad-hoc smoothed version. The smoothed version of $V$, $S$, $H$ keeps global chromological properties, meanwhile makes the network converge better, which is demonstrated in sub-figure (g).}
    
    \label{fig:hslmap}
    \end{figure}

    A straight forward solution to supervise $F_{val}, F_{sat}, F_{hue}$ is using the standard HSV decomposition equations to get HSV channels. However, we observe that this strategy could contain high frequency contents in the output channel as visualized in Fig.~\ref{fig:std-l}~$\sim$ Fig.~\ref{fig:smo-h}, which may degrade the convergence of network according to our experiments in Fig.~\ref{fig:loss_curve}.
    
    Therefore we heuristicly designed an approximated version of HSV loss to stabilize network training. 
    It is mainly based on a combination of several differentiable basic image processing filters (e.g. whitening, blurring, blending) to obtain smooth approximations of these three attributes $H,S,V$, which benifit training procedure.
    The implementation details are shown in supplementary.

    Auxiliary HSV losses $\mathcal{L}_{val}^{l}$, $\mathcal{L}_{sat}^{l}$, $\mathcal{L}_{hue}^{l}$ are calculated with MSE in low resolution stream only due to memory consideration. 
    We also apply total variation regularization on predicted filters to increase smoothness.
    The overall training loss is defined as follows, where $\lambda_i (i=1,...,5)$ is hyper-parameters:
    
    
    
    \begin{equation} \label{eq:eq_loss}
        \mathcal{L} = \lambda_1 \mathcal{L}_{rgb}^{l} + \lambda_2 \mathcal{L}_{rgb}^{h} + \lambda_3 \mathcal{L}_{val}^l + \lambda_4 \mathcal{L}_{sat}^l + \lambda_5 \mathcal{L}_{hue}^l
    \end{equation}
    
    \section{Experiments}

    \begin{table*}[!thb]
    
        \scriptsize
        \begin{center}
            \resizebox{\textwidth}{!}{
        \begin{tabular}{l|c|c|c|c|c|c|c|c|c|c}
        \toprule[1pt]
        
        \multirow{3}{*}{Method} & \multicolumn{2}{c|}{Entire Dataset} & \multicolumn{2}{c|}{HCOCO} & \multicolumn{2}{c|}{HAdobe5k} & \multicolumn{2}{c|}{HFlickr} & \multicolumn{2}{c}{Hday2night}  \\
        \cline{2-11}
        {}                     & MSE $\downarrow$  & PSNR $\uparrow$  & MSE $\downarrow$  & PSNR $\uparrow$  & MSE $\downarrow$  & PSNR $\uparrow$  & MSE $\downarrow$  & PSNR $\uparrow$  & MSE $\downarrow$  & PSNR $\uparrow$  \\
        %
        %
        %
        %
        %
        %
        %
        %
        \hline
        Input image 
        & 177.99 & 31.22 & 73.03 & 33.53 & 354.46 & 27.63 & 270.99 & 28.20 & 113.07 & 33.91 \\
        iDIH-HRNet\cite{sofiiuk2021foreground} 
        & - & - & 19.96 & 38.25 & - & - & 93.50 & 32.42 & 71.01 & 35.77   \\
        iDIH-HRNet\cite{sofiiuk2021foreground}+BU 
        & 43.56 & 34.98 & 34.40 & 35.45 & 37.82 & 35.47 & 104.69 & 30.91 & 50.87 & 37.41 \\ 
        iDIH-HRNet\cite{sofiiuk2021foreground}+GF\cite{hekaiming2013GF}   
        & 35.47 & 36.00 & 25.93 & 36.70 & 34.51 & 36.03 & 85.05 & 32.01 & \textbf{49.90} & \textbf{37.67} \\
        iDIH-HRNet\cite{sofiiuk2021foreground}+BGU\cite{chen2016bilateral}   
        & 26.85 & 37.24 & 18.53 & 37.90  & 26.71 & 37.50 & 66.26 & 33.19 & 51.96 & 37.23 \\
        DCCF 
        & \textbf{24.65} & \textbf{37.87}
        & \textbf{17.07} & \textbf{38.66} 
        & \textbf{23.34} & \textbf{37.75} 
        & \textbf{64.77} & \textbf{33.60} 
        & 55.76 & 37.40 \\
        \bottomrule[1pt]
        \end{tabular}}
        \end{center}
        \caption{\textbf{Quantative performance comparison} on the iHarmony4 test sets. 
        We are the first to evaluate on original resolution in this dataset. The best results are in bold. 
        We use recent state-of-the-art network iDIH-HRNet~\cite{sofiiuk2021foreground} and several post-upsampling methods as our baselines.
        '-' means not able to obtain results due to memory limitation.
        Our method is trained in an end-to-end manner and outperforms baselines by comparison. More quantative results with different backbones are shown in supplementary.
        }
        \vspace{-15pt}
        \label{tab:LR_results}
        \end{table*}

    \begin{figure*}[!thb]
        \centering

        \begin{subfigure}[t]{0.15\linewidth}
            \centering
            \includegraphics[width=1\linewidth]{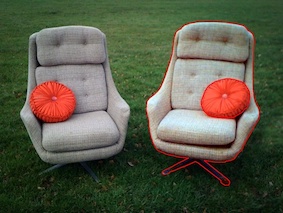}
        \end{subfigure}
        \begin{subfigure}[t]{0.15\linewidth}
            \centering
            \includegraphics[width=1\linewidth]{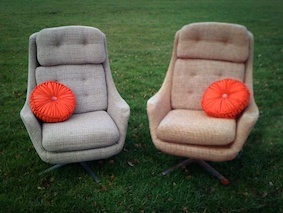}
        \end{subfigure}
        \begin{subfigure}[t]{0.15\linewidth}
            \centering
            \includegraphics[width=1\linewidth]{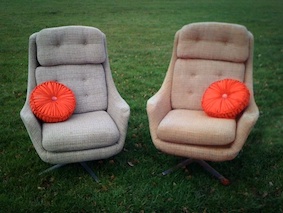}
        \end{subfigure}
        \begin{subfigure}[t]{0.15\linewidth}
            \centering
            \includegraphics[width=1\linewidth]{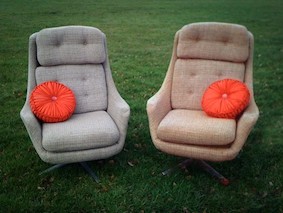}
        \end{subfigure}
        \begin{subfigure}[t]{0.15\linewidth}
            \centering
            \includegraphics[width=1\linewidth]{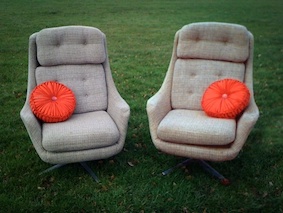}
        \end{subfigure}
        \begin{subfigure}[t]{0.15\linewidth}
            \centering
            \includegraphics[width=1\linewidth]{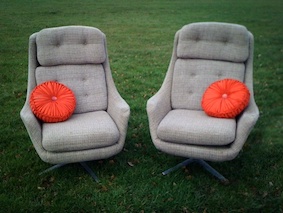}
        \end{subfigure}

        \begin{subfigure}[t]{0.15\linewidth}
            \centering
            \includegraphics[width=1\linewidth]{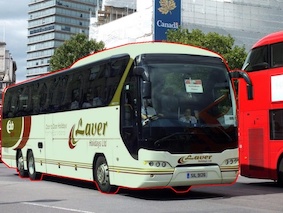}
        \end{subfigure}
        \begin{subfigure}[t]{0.15\linewidth}
            \centering
            \includegraphics[width=1\linewidth]{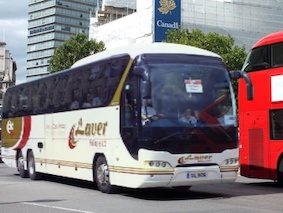}
        \end{subfigure}
        \begin{subfigure}[t]{0.15\linewidth}
            \centering
            \includegraphics[width=1\linewidth]{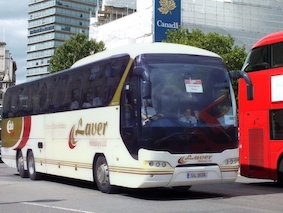}
        \end{subfigure}
        \begin{subfigure}[t]{0.15\linewidth}
            \centering
            \includegraphics[width=1\linewidth]{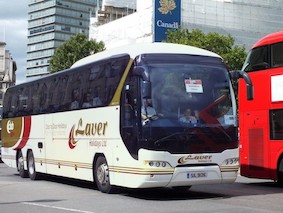}
        \end{subfigure}
        \begin{subfigure}[t]{0.15\linewidth}
            \centering
            \includegraphics[width=1\linewidth]{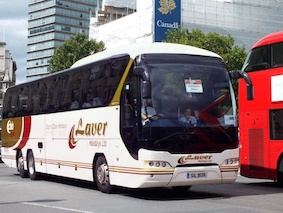}
        \end{subfigure}
        \begin{subfigure}[t]{0.15\linewidth}
            \centering
            \includegraphics[width=1\linewidth]{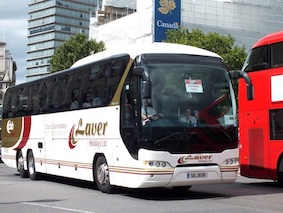}
        \end{subfigure}

        \begin{subfigure}[t]{0.15\linewidth}
            \centering
            \includegraphics[width=1\linewidth]{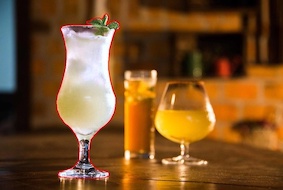}
        \end{subfigure}
        \begin{subfigure}[t]{0.15\linewidth}
            \centering
            \includegraphics[width=1\linewidth]{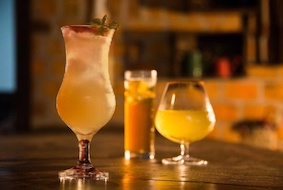}
        \end{subfigure}
        \begin{subfigure}[t]{0.15\linewidth}
            \centering
            \includegraphics[width=1\linewidth]{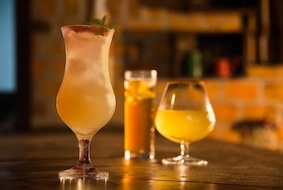}
        \end{subfigure}
        \begin{subfigure}[t]{0.15\linewidth}
            \centering
            \includegraphics[width=1\linewidth]{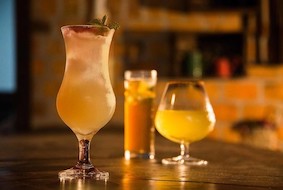}
        \end{subfigure}
        \begin{subfigure}[t]{0.15\linewidth}
            \centering
            \includegraphics[width=1\linewidth]{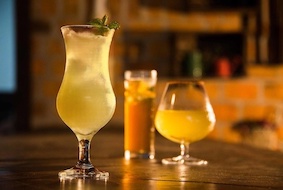}
        \end{subfigure}
        \begin{subfigure}[t]{0.15\linewidth}
            \centering
            \includegraphics[width=1\linewidth]{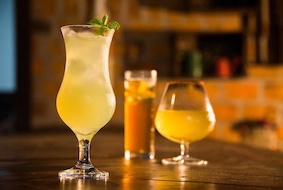}
        \end{subfigure}

        \begin{subfigure}[t]{0.15\linewidth}
            \centering
            \includegraphics[width=1\linewidth]{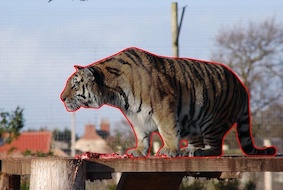}
        \end{subfigure}
        \begin{subfigure}[t]{0.15\linewidth}
            \centering
            \includegraphics[width=1\linewidth]{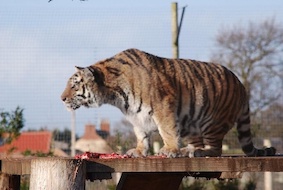}
        \end{subfigure}
        \begin{subfigure}[t]{0.15\linewidth}
            \centering
            \includegraphics[width=1\linewidth]{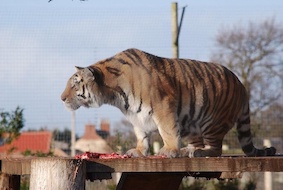}
        \end{subfigure}
        \begin{subfigure}[t]{0.15\linewidth}
            \centering
            \includegraphics[width=1\linewidth]{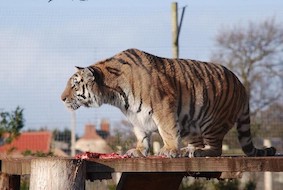}
        \end{subfigure}
        \begin{subfigure}[t]{0.15\linewidth}
            \centering
            \includegraphics[width=1\linewidth]{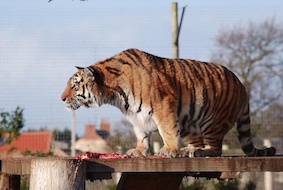}
        \end{subfigure}
        \begin{subfigure}[t]{0.15\linewidth}
            \centering
            \includegraphics[width=1\linewidth]{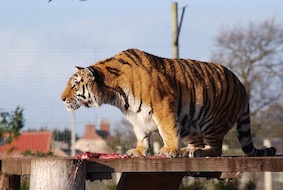}
        \end{subfigure}

        \begin{subfigure}[t]{0.15\linewidth}
            \centering
            \includegraphics[width=1\linewidth]{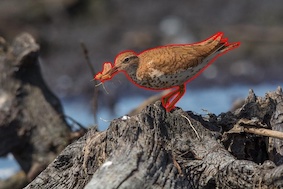}
            \caption{Input}
        \end{subfigure}
        \begin{subfigure}[t]{0.15\linewidth}
            \centering
            \includegraphics[width=1\linewidth]{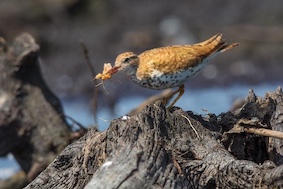}
            \caption{BU}
        \end{subfigure}
        \begin{subfigure}[t]{0.15\linewidth}
            \centering
            \includegraphics[width=1\linewidth]{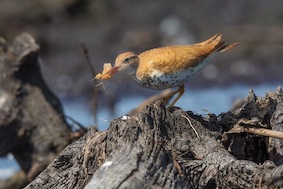}
            \caption{GF}
        \end{subfigure}
        \begin{subfigure}[t]{0.15\linewidth}
            \centering
            \includegraphics[width=1\linewidth]{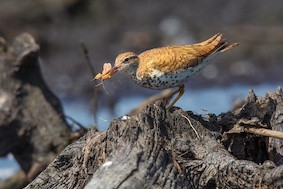}
            \caption{BGU}
        \end{subfigure}
        \begin{subfigure}[t]{0.15\linewidth}
            \centering
            \includegraphics[width=1\linewidth]{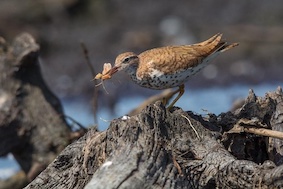}
            \caption{DCCF}
        \end{subfigure}
        \begin{subfigure}[t]{0.15\linewidth}
            \centering
            \includegraphics[width=1\linewidth]{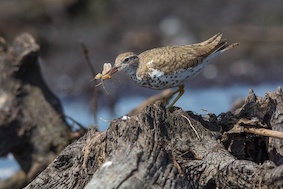}
            \caption{GT}
        \end{subfigure}
 
        \caption{\textbf{Visualization of high-resolution results}. Foregrounds are marked in red contour.
        Bilinear upsampling, guided filter upsampling and bilateral guided upsampling are represented as BU, GF~\cite{hekaiming2013GF} and BGU~\cite{chen2016bilateral} respectively. 
        GT represents ground truths. 
        Our method DCCF has not only better global appearance but also refined high resolution details.
        Zoom for better view.
        More visual results please refer to supplementary materials.}
        \vspace{-15pt}
        \label{fig:HR_vis}
        
    \end{figure*}


    In this section, we first describle experimental setups and implementation details, then compare our approach with the state-of-the-arts quantitatively and qualitatively. 
    Finally, we carry out some ablation studies and provide a simple comprehensible interface to interact with our model. 
    We also present more results and potential limitations in supplementary materials.

    \subsection{Experimental Setups}
    We use iHarmony4~\cite{cong2020dovenet} as our experiment dataset which contains 73146 images. 
    It consists of 4 subsets: HCOCO, HFlickr, HAdobe5k, HDay2night.
    The image resolution varies from 640$\times$480 to 6048$\times$4032,
    which is difficult for learning based color harmonization algorithms to process on the original images' full resolution.
    We suggest readers refer to~\cite{cong2020dovenet} for dataset details.
    
    Since the lack of high-resolution process ability, previous methods~\cite{tsai2017deep, cun2020improving, sofiiuk2021foreground, Ling_2021_CVPR, cong2020dovenet, cong2021bargainnet, guo2021intrinsic} resize all images in the dataset to 256$\times$256 to process and evaluate their performance via Mean Square Error (MSE) and Peak Signal To Noise Ratio (PSNR) in this extremely low-resolution.
    However, we argue that evaluate algorithms on the image's original full-resolution is much more scientific for practical applications.
    In this paper, we adopt MSE and PSNR as our objective metrics on the image's original full-resolution instead of 256$\times$256.

    \subsection{Implementation Details}
    Our DCCF learning framework is differentiable and could be stacked on the head of any deep feature extraction networks.
    In this paper, we adopt the recent state-of-the-art harmonization network iDIH-HRNet~\cite{sofiiuk2021foreground} as our backbone to carry out experiments.
    For feature extraction backbone, we downsample inputs (images and corresponding foreground mask) to 256$\times$256 following the previous deep harmonization models' common setting.
    For detailed training procedure and hyper-parameter setting, please refer to our official Pytorch~\cite{paszke2019pytorch} code\footnote{https://github.com/rockeyben/DCCF}.
    
    \subsection{Comparison with Baselines}
    In order to evalute the effectiveness of our proposed DCCF learning framework, we construct two kinds of baselines. 
    (1) Applying recent state-of-the-art methods directly on the original input images to get the full-resolution harmonized results.
    (2) Applying recent state-of-the-art methods on the low-resolution inputs ($256\times256$) to predict low-resolution harmonized images  and adopting variates of state-of-the-art post-processing methods to get the final full-resolution harmonized results.
    In this paper, we choose iDIH-HRNet~\cite{sofiiuk2021foreground} as the deep model provided by Sofiiuk et al.~\cite{sofiiuk2021foreground} and Bilinear Upsampling (BU), Guided Filter Upsampling~\cite{hekaiming2013GF} (GF), Bilateral Guided Upsampling~\cite{chen2016bilateral} (BGU) as post-processing methods.
    For fair comparison, we adopt the same low-resolution (i.e. $256\times256$) feature extractor as \cite{sofiiuk2021foreground} for our DCCF learning framework.
    The performance comparison is shown in Table~\ref{tab:LR_results}. Some harmonization results are shown in Fig.~\ref{fig:HR_vis}. For the comparison of efficiency metrics like inference time and memory usage, please refer to supplementary for details.

    The method of applying~\cite{sofiiuk2021foreground} directly on the full-resolution (first row in Table~\ref{tab:LR_results}) performs pooly.
    The principal reason is that~\cite{sofiiuk2021foreground} is designed and trained on the resolution of $256\times256$, directly applying this model in testing phase to the original image full-resolution would lead to serious feature misalignment.
    Moreover this strategy failed on HAdobe5k subset (max resolution: 6048$\times$4032) due to memory limitation.

    The method of applying post-processing after the low-resolutional prediction results from~\cite{sofiiuk2021foreground} with low-resolution inputs solves the memory problem.
    However, BU would lead to blurring effect, especially for high-resolution subset HAdobe5k, see Fig~\ref{fig:HR_vis}.
    Therefore, we adopt more advanced post-processing algorithms GF~\cite{hekaiming2013GF} and BGU~\cite{chen2016bilateral} that take original full-resolution image as detail guidance to mitigate the blurring effect from upsampling operation.
    Table ~\ref{tab:LR_results} shows that these upsampling methods outperform bilinear upsampling methods by a large margin and the best one BGU\cite{chen2016bilateral} achieves 26.85 on MSE and 37.24 on PSNR.
    However, the best performance of post-processing methods is behind our approach DCCF.
    Our approach achieves 24.65 on MSE and 37.87 on PSNR, $7.63\%$ and $1.69\%$ relative improvements on MSE and PSNR respectively compared with ~\cite{sofiiuk2021foreground}+ BGU~\cite{chen2016bilateral}.
    \subsection{Qualitative Results}

    We conduct two evaluations to compare the subjective visual quality of 
    DCCF with other methods, which is presented in Table~\ref{tab:qualitative_results}. First, we adopt LPIPS~\cite{zhang2018perceptual} to evalute visual perceptual similarity of harmonized image and ground truth reference. It computes the feature distance between two images and the lower score indicates better result. Second, we randomly select 20 images then present DCCF result with baseline results on the screen after shuffling, and ask 12 users to judge images' global appearance and detail texture then give scores from 1 to 5, the higher the better. Our DCCF achieves the best result in both metrics which is consistent the quantative performance.

    \begin{table}[!htb]
        \scriptsize
        \vspace{-15pt}
            \centering
            \resizebox{0.9\textwidth}{!}{
              \begin{tabular}{l|c|c|c|c}
              \toprule[1pt]
              {Method} & iDIH-HRNet\cite{sofiiuk2021foreground} +BU
               & iDIH-HRNet\cite{sofiiuk2021foreground}+GF~\cite{hekaiming2013GF} 
               & iDIH-HRNet\cite{sofiiuk2021foreground}+BGU\cite{chen2016bilateral} 
               & DCCF \\
            \hline
              LPIPS~\cite{zhang2018perceptual} $\downarrow$ & 0.0459 & 0.0291 & 0.0201 & \textbf{0.0186} \\
                    \hline
              User Score $\uparrow$  & 2.0541  & 2.6583 & 3.3041 
              & \textbf{3.5583} \\
              \bottomrule[1pt]
              \end{tabular}}
              \vspace{5pt}
              \caption{\textbf{Qualitative results}. We evaluate visual perceptual quality by DNN-based image quality accessment LPIPS~\cite{zhang2018perceptual} and a user study.}
              \label{tab:qualitative_results}
        \vspace{-20pt}
        
    \end{table}

    \subsection{Ablation Studies}

    In this subsection, we carry out a number of ablations to analyze our DCCF learning framework from the aspect of filter and loss function design.
    
    \textbf{Filter Design:} An evaluation of filter design is shown in Table~\ref{tab:ablation_loss}a.
    DBL~\cite{gharbi2017deep} is an end-to-end "black-box" bilateral learning method that proposed in image enhancement.
    We adapt it to our DCCF learning framework to process high-resolution image harmonization.
    DCCFs w.o. attention is our DCCF learning method that exclude attentive rendering filter.
    Even DCCFs w.o. attention improves the performance of DBL filter~\cite{gharbi2017deep} by 1.56 (5.58\%) on MSE.
    It demonstrate that the performance of our model is not just from end-to-end training, 
    our divide, conquer and assemble strategy that learns explicit meaningful parameters also benefit a lot for color harmonization task.
    DCCFs with attn further improve the DBL filter\cite{gharbi2017deep} by 3.27 (11.71\%) on MSE.
    
    \textbf{Loss Functions:} The impact of loss functions for our DCCF learning framework is shown in Table~\ref{tab:ablation_loss}b.
    Note that standard $H$ channel is an angle value while our approximated $H$ is a scalar value, so we train standard $\mathcal{L}_{h}$ with cosine distance while training approximated smooth $\mathcal{L}_{h}$ with euclidean distance.
    Numerical results show that supervisions from HSV color space is essential for our DCCF learning framework, which is manifested in simply adding loss from standard HSV channels will remarkably decrease MSE from 35.17 to 27.86.
    The principal reason may be the parameters of our DCCFs (expect for the last attentive rendering filter) are designed from the inspiration of practical tuning criteria in HSV color space used by color artists and has explicit chromatics meaning.
    Therefore model converges better when supervisory signals from HSV color space are added, which is demonstrate in Fig.~\ref{fig:loss_curve}.
    It is worth noting that adding smooth approximated HSV loss described in subsection~\ref{section:smooth} instead of standard HSV loss will further decrease MSE to 24.65 which demonstrates the effectiveness of proposed smoothing HSV loss.

    \begin{table}[!htb]
        \scriptsize
        \centering
        \vspace{-10pt}
        \begin{minipage}{.45\linewidth}
          \centering       
          \resizebox{1\textwidth}{!}{
            \begin{tabular}{l|c|c}
            \toprule[1pt]
            {Method} & MSE $\downarrow$  & PSNR $\uparrow$\\
            \hline
            DBL~\cite{gharbi2017deep}  & 27.92 & 37.48 \\
            DCCFs w.o. attention   & 26.36 & 37.80 \\
            DCCFs with attention  & \textbf{24.65} & \textbf{37.87} \\
            \bottomrule[1pt]
            \end{tabular}}
            \caption*{(a) Ablations for filter design}
        \end{minipage}%
        \begin{minipage}{.45\linewidth}
              \centering
              \resizebox{1\textwidth}{!}{
                \begin{tabular}{l|c|c}
                \toprule[1pt]
                {Method}
                         & MSE $\downarrow$  & PSNR $\uparrow$\\
                \hline
                $\mathcal{L}_{rgb}$  &
                 35.17  & 36.81 \\	
                $\mathcal{L}_{rgb}$ + standard $\mathcal{L}_{hsv}$   &
                  27.86 & 37.39  \\
                $\mathcal{L}_{rgb}$ + smooth $\mathcal{L}_{hsv}$ &
                 \textbf{24.65} & \textbf{37.87} \\
                \bottomrule[1pt]
                \end{tabular}}
                \caption*{(b) Ablations for loss function}
            \end{minipage} 
        \caption{
        \textbf{Ablation studies} on the iHarmony4 dataset. 
        (1) As for filter design, DBL~\cite{gharbi2017deep} is a "black-box" per-pixel linear filter that directly applied to RGB images. DCCFs with attention achieves the best result. 
        (2) As for losses, supervisions from HSV is essential for DCCF learning framework.
        Supervisions we constructed from HSV (i.e. smooth $\mathcal{L}_{hsv}$) improves standard HSV by 3.21 (11.65\%) on MSE.
        }
        \vspace{-30pt}
        \label{tab:ablation_loss}
    \end{table}

    \subsection{Comprehensible Interaction with Deep Model}
    
    \begin{figure}[!thb]
    \centering
    \includegraphics[width=0.8\textwidth]{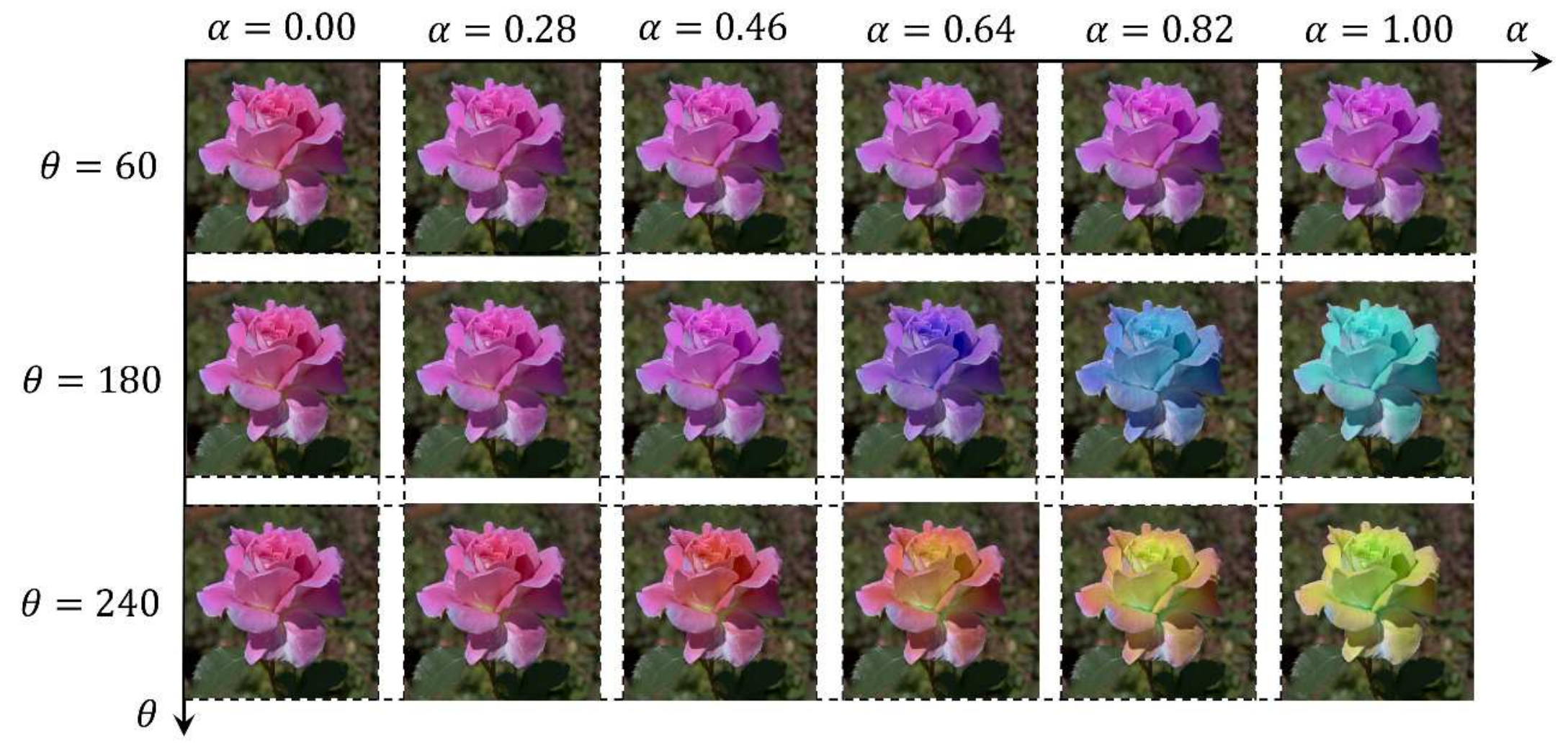}
    \caption{Illustration of comprehensible interaction with deep harmonization model on parameter space of hue adjustment.
    Abscissa represents parameter $\alpha$ and ordinate represents parameter $\theta$.
    Sampling values in $(\alpha, \theta)$ and their results are listed.
    Zoom for better view.
    }
    \vspace{-20pt}
    \label{fig:edit_filter}
    \end{figure}
    
     Benefiting from the comprehensible neural filters, we could provide a simple yet efficient handler for users to cooperate with deep model to get the desired results with very little effort when necessary.
    We provide two adjustable parameters 
    in the three dimensions of hue, saturation and value respectively for users to express their color adjustment intentions.
    For space limitation, we only explain hue adjustment for example.
    The other two dimensions are similar and will be detailed in supplementary.
    
    For hue, we define parameter $\theta \in [0,360]$ and $\alpha \in [0,1]$ to represent the angle for Hue circle and the amount of user color intentions respectively.
    We calculate the desired rotation matrix $R$ mentioned in Eq. (\ref{eq:hue_filter}) as:

    \begin{equation}
    \label{eq:rotation}
    \begin{bmatrix}
    \frac{1}{3}-\frac{2\cos\theta}{3} & \frac{1-\cos\theta}{3} - \frac{\sin\theta}{\sqrt{3}} & 
    \frac{1-\cos\theta}{3} + \frac{\sin\theta}{\sqrt{3}}
    \\
    \frac{1-\cos\theta}{3} + \frac{\sin\theta}{\sqrt{3}} & \frac{1}{3}-\frac{2\cos\theta}{3} & 
    \frac{1-\cos\theta}{3} - \frac{\sin\theta}{\sqrt{3}}
    \\
    \frac{1-\cos\theta}{3} + \frac{\sin\theta}{\sqrt{3}} & \frac{1-\cos\theta}{3} + \frac{\sin\theta}{\sqrt{3}} & \frac{1}{3}-\frac{2\cos\theta}{3} 
    \end{bmatrix}
    \end{equation}
    
    Then we could get the final rotation matrix $R$: $F_{hue}'=\alpha * R + (1-\alpha) * F_{hue}$, which can be applied on image that takes global user intentions and local complex self-adaptions from deep model in mind.
    
    In one word, users can express their color intentions by parameter $\theta \in [0,360]$ and decide the amount of injected color by controlling $\alpha \in [0,1]$, which is illustrated in Fig.~\ref{fig:edit_filter}.
    It is worth noting that when users interact with deep model in one dimension (such as hue above), they need not worry about the side-effect changes of other two dimensions from the network's prediction.



    
    \section{Conclusion}
    
    In this paper, we propose comprehensible image processing filters to deal with image harmonization problem. 
    By gradually modifying image's attributes: value, saturation and hue, we can obtain results not only high-quality but also understandable. 
    This also facilitate human to cooperate with deep models to perform image harmonization.
    We also leverage these filters to tackle high resolution images in a simple yet effective way. 
    We hope that DCCF can set up a brand new direction for image harmonization.

\clearpage

%
%
\bibliographystyle{splncs04}
\bibliography{egbib}

\title{Supplementary Material for DCCF: Deep Comprehensible Color Filter Learning Framework for High-Resolution Image Harmonization} 
\titlerunning{DCCF: Deep Comprehensible Color Filter Learning Framework}
%
\author{Ben Xue\inst{1}\thanks{Finish this work during an internship at Alibaba Group. Email: \email{xueben@pku.edu.cn}.}\and
Shenghui Ran\inst{2} \and
Quan Chen\inst{2}\thanks{Corresponding author. Email: \email{myctllmail@163.com}. }\and \\
Rongfei Jia\inst{2}\and
Binqiang Zhao\inst{2}\and
Xing Tang\inst{2}\\
}

%
\institute{
Academy for Advanced Interdisciplinary Studies, Peking University, China \and
Alibaba Group, China}

\maketitle

\section{Introduction}

For better acknowledgement of details, we provide several supplementary sections to demonstrate the mechanism and capability of DCCF.

In Section \ref{sec:high-res-result} and \ref{sec:high-res-efficiency}, we provide more experiment results on high-resolution images along with the efficiency comparison.
In Section \ref{sec:low-res-results}, we further show the low-resolution results to demonstrate the efficacy of designed filters. 

In Section \ref{color filter}, we show that a constrained rotation matrix in RGB space can actually rotate the color angle in HSV space. 
Along with the disentanglement in \textit{high resolution assembly module} in Section \ref{disentangle}, we also show that DCCF like $F_{hue}$ can learn unconstrained rotation matrix which can be projected on its current HSV plane to directly affect the specific HSV channel.

In Section \ref{sec:auxiliary-hsv-loss}, we demonstrate the numerical procedure of standard and smoothing(ours) HSV supervision strategies.

In Section \ref{intermediate}, we show the capability of DCCF to adjust specific image attribute. 
Each filter of this family can be picked out to qualify its sub-task.
In Section \ref{interaction} and Section \ref{sec:high-res-vis}, we provide more visualization results of comprehensible interaction and high-resolution results.
In Section \ref{sec:limitation}, we discuss the potential limitation of our framework.

\section{High-Resolution Results with Different Backbones}
\label{sec:high-res-result}

Note that our DCCF can be plugged into different backbones, we select DIH~\cite{tsai2017deep} and S$^2$AM~\cite{cun2020improving} for experiment. 
~\cite{tsai2017deep} shares similary U-net\cite{Unet} architecture with iDIH-HRNet~\cite{sofiiuk2021foreground} except for the extra pretrained visual feature from HRNet~\cite{SunXLW19}. 
~\cite{cun2020improving} uses spatial-separated attention module in decoder to aggregate semantic information. 
For fair comparison, we keep learning strategy the same as DCCF when training these backbones, where we use RandomResizedCrop and RandomHorizontalFlip as data augmentation, foreground-normlized MSE~\cite{sofiiuk2021foreground} as training loss, XavierGluon (gaussian, magnitute=0.2) as weights initializer. 
We use bilinear upsamplng (BU), guided filter (GF~\cite{hekaiming2013GF}) and bilateral grid upsampling (BGU~\cite{chen2016bilateral}) as post-processing methods.
Experiments in Table~\ref{tab:supp_hr_result} show that DCCF constantly outputs these baselines which demonstrates the robustness and generality of our framework.

\begin{table}[!thb]
   \scriptsize
   \begin{center}
    \resizebox{\textwidth}{!}{
   \begin{tabular}{l|c|c|c|c|c|c|c|c|c|c}
   \toprule[1pt]
   \multirow{3}{*}{Method} & \multicolumn{2}{c|}{Entire Dataset} & \multicolumn{2}{c|}{HCOCO} & \multicolumn{2}{c|}{HAdobe5k} & \multicolumn{2}{c|}{HFlickr} & \multicolumn{2}{c}{Hday2night} \\
   \cline{2-11}
   {}  & MSE $\downarrow$  & PSNR $\uparrow$   & MSE $\downarrow$  & PSNR $\uparrow$ & MSE $\downarrow$  & PSNR $\uparrow$ & MSE $\downarrow$  & PSNR $\uparrow$ & MSE $\downarrow$  & PSNR $\uparrow$ \\
   \hline
   
   DIH~\cite{tsai2017deep}
   & - & - & 36.39 & 36.56 & - & - & 186.38 & 30.78 & 61.89 & 35.40 \\
   DIH~\cite{tsai2017deep} + BU
   & 51.28 & 34.39 & 39.35 & 34.92 & 44.99 & 34.81 & 129.48 & 30.14 & 51.07 & 36.77 \\
   DIH~\cite{tsai2017deep} + GF\cite{hekaiming2013GF}
   & 43.10 & 35.35 & 30.73 & 36.11 & 41.57 & 35.34 & 109.99 & 31.13 & 50.00 & 37.09 \\
   DIH~\cite{tsai2017deep} + BGU\cite{chen2016bilateral}
   & 34.37 & 36.47 & 23.16 & 37.21 & 34.11 & 36.63 & 90.24 & 32.18 & 51.78 & 36.64 \\
   DCCF-DIH 
   & \textbf{33.39} & \textbf{36.87} & \textbf{21.60} & \textbf{37.81} & \textbf{34.09} & \textbf{36.77} & \textbf{89.86} & \textbf{32.24} & \textbf{49.93} & \textbf{37.23} \\
   \hline
   S$^2$AM~\cite{cun2020improving}
   & - & - & 33.43 & 36.93	& - & - & 186.70 & 30.78 & 57.48 & 36.05 \\
   S$^2$AM~\cite{cun2020improving} + BU
   & 44.02 & 35.02 & 36.36 & 35.30 & 33.01 & 35.95 & 112.74 & 30.77 & 41.84 & 37.44 \\
   S$^2$AM~\cite{cun2020improving} + GF\cite{hekaiming2013GF}
   & 35.88 & 36.05 & 27.83 & 36.53 & 29.63 & 36.58 & 93.06 & 31.85 & \textbf{40.94} & \textbf{37.79} \\                     
   S$^2$AM~\cite{cun2020improving} + BGU\cite{chen2016bilateral}
   & 27.94 & 37.18 & 20.25 & 37.73 & 24.71 & \textbf{37.68} & 73.82 & 32.99 & 42.49 & 37.28 \\
   DCCF-S$^2$AM 
   & \textbf{26.74} & \textbf{37.59} & \textbf{18.41} & \textbf{38.43} & \textbf{24.39} & 37.65 & \textbf{73.18} & \textbf{33.12} & 44.25 & 37.36 \\
   \bottomrule[1pt]
   \end{tabular}}
   \end{center}

   \caption{\textbf{Quantative results} on the iHarmony4 original-resolution test sets with other backbones. '-' means not able to obtain results due to memory limitation. 'DCCF-*' means DCCF filters backboned by *.}
   \label{tab:supp_hr_result}

   \vspace{-20pt}
   \end{table}

\section{Comparison with CDTNet}

We finetune our model under weaker resolution settings (1024$\times$1024, 2048$\times$2048) on HAdobe5k subset to compare with the recent high resolution harmonization method CDTNet~\cite{CDTNet}. To make fair comparison, we use the same backbone S$^2$AM-256 with ~\cite{CDTNet} (i.e CDTNet-256). The result is shown in Table \ref{tab:cdtnet}. Our DCCF has better performance on higher resolution setting (2048$\times$2048) than CDTNet-256. It is also observed that the performance of ~\cite{CDTNet} drops significantly as resolution increases, while our method maintains stable performance. Note that other high resolution experiments in our paper are conducted under a much stronger setting: the original resolution of HAdobe5k can range up to 6048$\times$4032.

   \begin{table}[!htb]
        \scriptsize
            \centering
            \vspace{-5pt}
            \resizebox{0.6\textwidth}{!}{
              \begin{tabular}{l|l|c|c|c|c}
               \toprule[1pt]
              Resolution & {Method} & MSE$\downarrow$ & PSNR$\uparrow$ & fMSE$\downarrow$ & SSIM$\uparrow$ \\
              \hline 
              1024 & CDTNet-256 & 21.24 & \textbf{38.77} & \textbf{152.13} & \textbf{0.9868}  \\
               $\times$1024& DCCF & \textbf{21.12} & 38.38 & 171.17 & 0.9852 \\
               \hline
              2048 &CDTNet-256 & 29.02 & 37.66 & 198.85 & 0.9845\\
              $\times$2048 &DCCF & \textbf{21.35}	& \textbf{38.47} & \textbf{174.78} & \textbf{0.9856} \\
              \bottomrule[1pt]
              \end{tabular}}
              \vspace{7pt}
              \caption{\textbf{Quantative comparison} with CDTNet~\cite{CDTNet} on HAdobe5k subset.}
              \label{tab:cdtnet}
        
    \end{table}

\section{Efficiency Comparison}
\label{sec:high-res-efficiency}

To compare efficiency between DCCF and existed post-processing upsampling methods, we test these methods on different resolutions from 1024$\times$1024 to 3072$\times$3072.
The experiment is conducted on a x86-64 machine (72 cores, ubuntu 18.04) with a 12GB Nivida Titan X gpu card.
We test cpu time (T-C), gpu time (T-G) and memory usage (Mem) for evaluation metrics.
Each method is warmed up by 10 times and averaged by another 20 times forward passes.
As for detailed parameters which could influence efficiency metrics, GF~\cite{hekaiming2013GF} uses $r=8$ kernel size, BGU~\cite{chen2016bilateral} uses default $16\times16\times3\times4$ grid size.

Experiments in Table~\ref{tab:efficiency} show that GF~\cite{hekaiming2013GF} take most of the leads in efficiency metrics, that is mainly because it only involves several basic box filters, however it losses too much performance compared to DCCF (MSE/PSNR, 24.65/37.87$\rightarrow$35.47/36.00). 
BGU~\cite{chen2016bilateral} explicitly estimates bilateral grids which contain image-to-image transformation coefficients, therefore the performance is higher compared to GF~\cite{hekaiming2013GF}, however the efficiency drops far behind since it needs extra optimization procedure.
To this end, DCCF achieves a good trade-off between performance and efficiency.

\begin{table*}[!thb]
        \scriptsize
        \begin{center}
            \resizebox{\textwidth}{!}{
        \begin{tabular}{l|c|c|c|c|c|c|c|c|c}
        \toprule[1pt]
        
        \multirow{3}{*}{Method} & \multicolumn{3}{c|}{$1024\times1024$} & \multicolumn{3}{c|}{$2048\times2048$} & \multicolumn{3}{|c}{$3072\times3072$}  \\
        \cline{2-10}
        {}  & T-C (ms) $\downarrow$  & T-G (ms) $\downarrow$  & Mem (MB) $\downarrow$  
        & T-C (ms) $\downarrow$  & T-G (ms) $\downarrow$  & Mem (MB) $\downarrow$
        & T-C (ms) $\downarrow$  & T-G (ms) $\downarrow$  & Mem (MB) $\downarrow$ \\
        \hline
        iDIH-HRNet\cite{sofiiuk2021foreground} 
        & 420  & 231 & 1641 
        & 41040 & 907 & 4233 
        & 139768 & 2042 & 8551 \\
        iDIH-HRNet\cite{sofiiuk2021foreground} + GF\cite{hekaiming2013GF}  
        & 642 & 80.2 & 983 
        & 2001 & 160  & 1513 
        & 10181 & 391 & 2483 \\
        iDIH-HRNet\cite{sofiiuk2021foreground} + BGU\cite{chen2016bilateral} 
        & 9932 & - & 2893
        & 20803 & - & 4042
        & 29836 & - & 8173 \\
        DCCF-iDIH-HRNet 
        & 762 & 104 & 1259 
        & 3289 & 286 & 2607
        & 6517 & 545 & 4845 \\

        \bottomrule[1pt]
        \end{tabular}}
        \end{center}
        \caption{\textbf{Efficiency comparison} between DCCF and advanced post-processing methods on different resolutions from 1024 to 3072. 
        'T-C' represents cpu time (ms), 'T-G' represents gpu time (ms) and 'Mem' represents memory usage (MB).
        Note that '-' means no offical implementation is found.
        }
        \vspace{-20pt}
        \label{tab:efficiency}
\end{table*}

\section{Low-Resolution Results}
\label{sec:low-res-results}

To further show the efficacy of our designed comprehensible filters, we compare our approach with other state-of-the-art deep models [27,6,5,9] on the iHarmony4 low-resolution ($256\times256$) test sets. 
The results are shown in Table~\ref{tab:supp_lr_results}.
It is interesting that our approach also outperforms the previous best one~ [23] slightly on the entire iHarmony4 dataset on low-resolution, which may due to the appropriate design of filters and extra supervision from auxiliary HSV losses via our framework.

\begin{table}[!thb]
   \scriptsize
   \begin{center}
    \resizebox{\textwidth}{!}{
   \begin{tabular}{l|c|c|c|c|c|c|c|c|c|c}
   \toprule[1pt]
   \multirow{3}{*}{Method} & \multicolumn{2}{c|}{Entire Dataset} & \multicolumn{2}{c|}{HCOCO} & \multicolumn{2}{c|}{HAdobe5k} & \multicolumn{2}{c|}{HFlickr} & \multicolumn{2}{c}{Hday2night} \\
   \cline{2-11}
   {}  & MSE $\downarrow$  & PSNR $\uparrow$   & MSE $\downarrow$  & PSNR $\uparrow$ & MSE $\downarrow$  & PSNR $\uparrow$ & MSE $\downarrow$  & PSNR $\uparrow$ & MSE $\downarrow$  & PSNR $\uparrow$ \\
   \hline
   
   DIH~\cite{tsai2017deep}
                           & 76.77 & 33.41 & 51.85 & 34.69 & 92.65 & 32.28 & 163.38 & 29.55 & 82.34 & 34.62 \\
   
   S$^2$AM~\cite{cun2020improving}
                           & 59.67 & 34.35 & 41.07 & 35.47 & 63.40 & 33.77 & 143.45 & 30.03 & 76.61 & 34.50  \\
                          
   DoveNet~\cite{cong2020dovenet}
                           & 52.36 & 34.75 & 36.72 & 35.83 & 52.32 & 34.34 & 133.14 & 30.21 & 54.05 & 35.18  \\
   
   IntrinsicIH~\cite{guo2021intrinsic} & 38.71 & 35.90 & 24.92 & 37.16 & 43.02 & 35.20 & 105.13 & 31.34 & 55.53 & 35.96 \\
   
   iDIH-HRNet~\cite{sofiiuk2021foreground}
                        & 22.81 & 38.18 & \textbf{14.35} & \textbf{39.53} & 23.43 & 37.18 & 61.42 & 33.84 & \textbf{45.09} & \textbf{38.08} \\

   DCCF
                        & \textbf{22.05} & \textbf{38.50} & 14.87 & 39.52 & \textbf{19.90} & \textbf{38.27} & \textbf{60.41} & \textbf{33.94} & 49.32 & 37.88 \\
   \bottomrule[1pt]
   \end{tabular}}
   \end{center}

   \caption{\textbf{Quantative results} on the iHarmony4 low-resolution test sets.}
   \vspace{-20pt}
   \label{tab:supp_lr_results}
   \end{table}

   \section{Ablation Study on Operation Order}

   According to our survey, many designers and artists in Photoshop community tend to harmonize an image in the order of 'value, saturation, hue'. We regard this phenomenon as a common convention thus design our framework in such an order.
   However we think the operation order of DCCF filter is also meaningful thus we make corresponding ablation experiment.
   We investigate the whole $3!$ combinations
   'VSH', 'HVS', 'SHV', 'VHS', 'HSV', 'SVH' 
   on low resolution images in Table \ref{tab:ablation_order}. Note that the results are slightly different with Table~\ref{tab:supp_lr_results} because Table~\ref{tab:ablation_order} uses fewer training epochs, but we ensure that all abaltions in Table~\ref{tab:ablation_order} share the same parameter setting. 
   A tentative conclusion is that the order would affect final results and different subsets also require different optimal operation orders. 
   A possible solution is introducing re-enforcement learning (RL) to decide which is the best operation order when processing a certain given image. This may inspire our future works.

       \begin{table}[!htb]
           \scriptsize
               \centering
               \resizebox{\textwidth}{!}{
                 \begin{tabular}{l|c|c|c|c|c|c|c|c|c|c}
                 \toprule[1pt]
                 \multirow{3}{*}{Operation Order} & \multicolumn{2}{c|}{Entire Dataset} & \multicolumn{2}{c|}{HCOCO} & \multicolumn{2}{c|}{HAdobe5k} & \multicolumn{2}{c|}{HFlickr} & \multicolumn{2}{c}{Hday2night} \\
                 \cline{2-11}
                 {}  & MSE $\downarrow$  & PSNR $\uparrow$   & MSE $\downarrow$  & PSNR $\uparrow$ & MSE $\downarrow$  & PSNR $\uparrow$ & MSE $\downarrow$  & PSNR $\uparrow$ & MSE $\downarrow$  & PSNR $\uparrow$ \\
                 \hline
                 V$\rightarrow$S$\rightarrow$H (default) & 22.52 & 38.57 & 14.20 & \textbf{39.73} & 22.54 & 38.12 & 61.80 & 33.92 & 45.54 & 37.51 \\
                 H$\rightarrow$V$\rightarrow$S & 22.90 & 38.53 & 14.45 & 39.66 & 23.66 & 38.12 & 60.32 & 33.93 & 49.78 & 37.67 \\
                 S$\rightarrow$H$\rightarrow$V & 22.88 & 38.43 & 14.90 & 39.48 & 21.81 & 38.17 & 63.94 & 33.74 & \textbf{41.64} & \textbf{37.94} \\
                 V$\rightarrow$H$\rightarrow$S & \textbf{22.11} & \textbf{38.63} & \textbf{14.02} & 39.71 & 21.98 & 38.35 & \textbf{59.54} & \textbf{33.98} & 51.77 & 37.58\\
                 H$\rightarrow$S$\rightarrow$V & 22.45 & 38.57 & 14.67 & 39.57 & \textbf{21.29} & \textbf{38.45} & 61.97 & 33.87 & 45.78 & 37.69 \\
                 S$\rightarrow$V$\rightarrow$H & 22.94 & 38.53 & 14.25 & 39.69 & 23.08 & 38.06 & 61.77 & 33.90 & 59.10 & 37.33 \\
                 \bottomrule[1pt]
                 \end{tabular}}
                 \vspace{7pt}
                 \caption{\textbf{Ablation study} of different orders.}
                 \label{tab:ablation_order}
           \vspace{-10pt}
       \end{table}
   
   \section{Hue Filter and Disentanglement}
   \label{sec:hue-filter-disentangle}
  
   \subsection{Hue Filter}
   \label{color filter}
   Let $\bm{x}:(x_r,x_g,x_b)$ indicates the RGB values for one pixel in image, $\bm{z}:(z_h,z_s,z_l)$ is the corresponding point in HSV space. Let $\bm{\Delta}$ is a learnable 3x4 affine transformation matrix in RGB space that contains a rotation matrix $\bm{R}$ and translation vector $t$, and $\bm{r}$ is a radian moving on the hue ring in HSV space.
   
\begin{figure}[htbp]
   \centering
   
   \begin{subfigure}[t]{0.3\linewidth}
   \centering
   \includegraphics[width=1\linewidth]{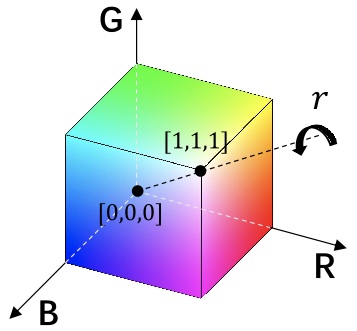}
   \caption{3D Rotation}
   \end{subfigure}
   \begin{subfigure}[t]{0.23\linewidth}
   \centering
   \includegraphics[width=1\linewidth]{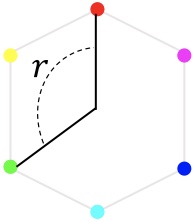}
   \caption{Diagonal}
   \end{subfigure}
   \begin{subfigure}[t]{0.25\linewidth}
   \centering
   \includegraphics[width=1\linewidth]{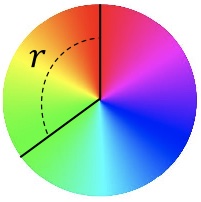}
   \caption{Hue}
   \end{subfigure}
   
   \caption{Illustration of rotation matrix. Viewing 3D RGB cube model (a) from the diagonal perspective, we can get the hexagon in (b), which is an approximation of real color circle in hue space (c).
   Therefore rotation $\bm{r}$ is equivalent in above three models.
   }
   \label{fig:rotation}
   \end{figure}

   According to the theory in \cite{haeberli1993matrix}, to rotate the hue by $\bm{r}$, we perform a 3D rotation of RGB colors about the diagonal vector [1.0 1.0 1.0] as as illustrated in Fig.~\ref{fig:rotation}.
   The resulting matrix will rotate the hue of the input RGB colors. A rotation of $2\pi/3$ will exactly map Red into Green, Green into Blue and Blue into Red.
   The matrix processing in \cite{haeberli1993matrix} makes an approximation that the diagonal axis in RGB space is equivalent to the hue axis in HSV space.
   This transformation has one problem, however, the luminance of the input colors is not preserved. 
   This can be fixed by shearing the value plane to make it horizontal.

   %

   We suppose that one could find a suitable rotation matrix $\bm{R}$ in RGB color space that is equivalent to \cite{haeberli1993matrix}.
   Therefore, it is possible to learn an affine color transformation function $f_{hue}(\bm{x}; \bm{\Delta})$ in RGB color space, which contains a rotation function $\bm{R}$ that could be the parameters for the corresponding hue rotation function $f_{hue}(h; \bm{r})$ in HSV space.
   Exactly, $f_{hue}(h; \bm{r})$ is the desired linear transformation for hue filter $F_{hue}$.
   It is obvious that the linear transformation $\bm{\Delta}$ could map one RGB point $\bm{x_1}$ to any other RGB point $\bm{x_2}$, and the corresponding HSV point $\bm{z_1}$ moves to $\bm{z_2}$.
   To avoid the modification along L and S axis, we perform HSV disentangle by projecting the path $(\bm{z_1} \rightarrow \bm{z_2})$ on H plane to get $\bm{z}_{2\parallel h}$.

   \subsection{Effect of HSV Disentanglement}
   \label{disentangle}
   
   \begin{figure}[htbp]
   \centering
   
   \begin{subfigure}[t]{0.3\linewidth}
   \centering
   \includegraphics[width=1\linewidth]{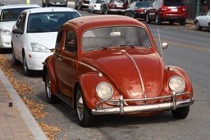}
   \caption{Input}
   \end{subfigure}
   \begin{subfigure}[t]{0.3\linewidth}
   \centering
   \includegraphics[width=1\linewidth]{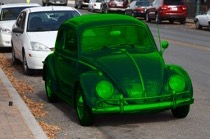}
   \caption{$F_{hue}$ on RGB}
   \end{subfigure}
   \begin{subfigure}[t]{0.3\linewidth}
   \centering
   \includegraphics[width=1\linewidth]{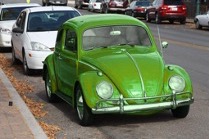}
   \caption{Disentangle}
   \end{subfigure}
   
   \caption{Illustration of disentanglement.
   \textit{High resolution assembly module} extracts the H channel of output and concatenates it with the input's V and S. This operation prevents $F_{hue}$ from changing V and S.
   }
   \label{fig:disentangle}
   \end{figure}
   
   We show the importance of \textit{high resolution assembly module} disentanglement in Fig.~\ref{fig:disentangle}.
   Taking $F_{hue}$ as example, naively applying it in RGB space actually changes V and S simultaneously. 
   DE (Disentangle) avoids this situation by extracting the H channel of output and concatenating it with the input's V and S. This ensures the action of $F_{hue}$ won't corrupt the result of $F_{val}$ and $F_{sat}$.
   
   Note that this disentanglement strategy is equivalent to the projection action in Section \ref{color filter}.
   We only need the path $(\bm{z_1} \rightarrow \bm{z_2})$ on the H plane: $(\bm{z}_{1\parallel h} \rightarrow \bm{z}_{2\parallel h} )$.
   Similarly, applying $F_{val}$ + DE and $F_{sat}$ + DE will generate projection path on V plane and S plane, which are $(\bm{z}_{1\parallel l} \rightarrow \bm{z}_{2\parallel l} )$, $(\bm{z}_{1\parallel s} \rightarrow \bm{z}_{2\parallel s} )$. The orthogonality of HSV space will ensure that these paths won't cross each other.



\section{Auxiliary HSV Loss}
\label{sec:auxiliary-hsv-loss}

\subsection{Standard HSV Decomposition}
\label{standard hsl}
The numerical conversion of HSV and RGB value is performed as:
\begin{equation}
	V = C_{max}
\end{equation}
\begin{equation}\label{eq:naive_s}
	S = \frac{C_{max} - C_{min}}{C_{max}}
\end{equation}
\begin{equation}\label{eq:naive_h}
H=
\left\{
\begin{array}{lr}
\pi/6 \times(\frac{G-B}{C_{max}-C_{min}} mod6), C_{max}=R &  \\
\pi/6 \times(\frac{B-R}{C_{max}-C_{min}} +2), C_{max}=G &  \\
\pi/6 \times(\frac{R-G}{C_{max}-C_{min}} +4), C_{max}=B &  \\
\end{array}
\right.
\end{equation}

This expression tends to result in noise points because it is based on numerical values rather than physical characteristics.

\subsection{Smoothing V}

The generation of $V_{smooth}$ is straightforward.
We apply gaussian blur on the original $V$ decomposition to get $V_{smooth}$, where we set variance scale $std=1.5$ and kernel size $K=5$ in implementation.

\subsection{Smoothing S}

\begin{figure}[htbp]
\centering

\begin{subfigure}[t]{0.17\linewidth}
\centering
\includegraphics[width=1\linewidth]{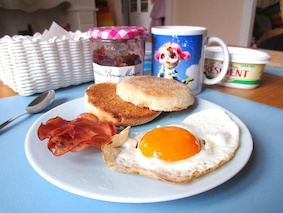}
\caption{Input}
\end{subfigure}
\begin{subfigure}[t]{0.17\linewidth}
\centering
\includegraphics[width=1\linewidth]{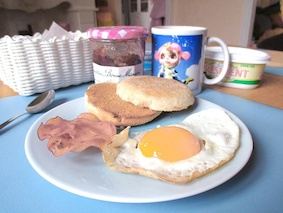}
\caption{Red $\downarrow$}
\end{subfigure}
\begin{subfigure}[t]{0.17\linewidth}
\centering
\includegraphics[width=1\linewidth]{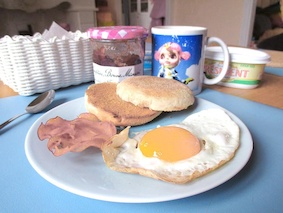}
\caption{Green $\downarrow$}
\end{subfigure}
\begin{subfigure}[t]{0.17\linewidth}
\centering
\includegraphics[width=1\linewidth]{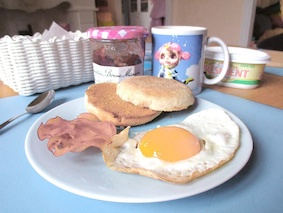}
\caption{Blue $\downarrow$}
\end{subfigure}
\begin{subfigure}[t]{0.17\linewidth}
\centering
\includegraphics[width=1\linewidth]{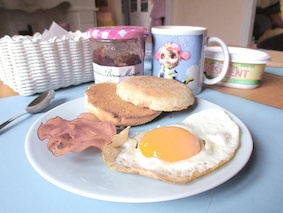}
\caption{Cyan $\downarrow$}
\end{subfigure}

\begin{subfigure}[t]{0.17\linewidth}
\centering
\includegraphics[width=1\linewidth]{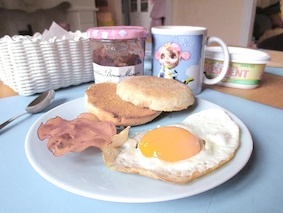}
\caption{Megenta $\downarrow$}
\end{subfigure}
\begin{subfigure}[t]{0.17\linewidth}
\centering
\includegraphics[width=1\linewidth]{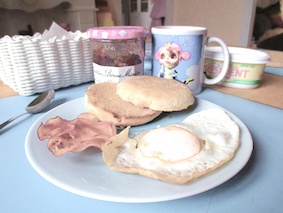}
\caption{Yellow $\downarrow$}
\end{subfigure}
\begin{subfigure}[t]{0.17\linewidth}
\centering
\includegraphics[width=1\linewidth]{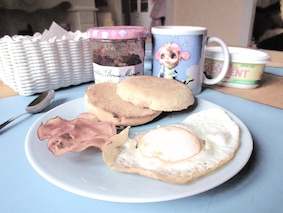}
\caption{Black $\uparrow$}
\end{subfigure}
\begin{subfigure}[t]{0.17\linewidth}
\centering
\includegraphics[width=1\linewidth]{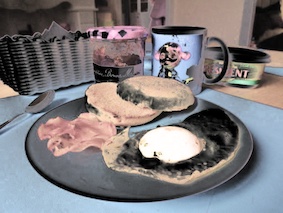}
\caption{White $\uparrow$}
\end{subfigure}
\begin{subfigure}[t]{0.17\linewidth}
\centering
\includegraphics[width=1\linewidth]{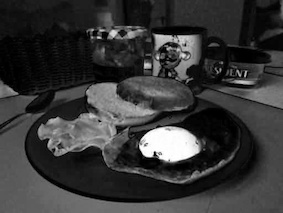}
\caption{Neutral $\uparrow$}
\end{subfigure}
\caption{Procedure of smoothing saturation map. It is arranged from left to right, top to bottom. Note that (j) is also the final result of our smoothing S.}

\label{fig:smap}
\end{figure}

Different from Eq. \ref{eq:naive_s}, we follow the operation in photoshop to get a smooth map.
First, we perform a selective color adjustment by setting all the colored tunes to -100\%: red, yellow, green, cyan, blue, and magenta(RGB, CMY).
Then for the blacks, whites and neutrals(BWN), we enhance them to 100\%. Noted that the param used here ranges from -100\% to 100\%, actually it is exactly the same $\sigma$ we used in our saturation filter.
The detailed procedure is shown in Fig.~\ref{fig:smap}.
The numerical expression can be expressed as:
\begin{align}
	&S_{t} = F_{sat}(S_{t-1};\sigma_t) * ROI(c_t), t=1,2,...9 \\
	&S_{smooth} = I_9, S_0 = I_1 = F_{val}(I)
\end{align}

Where $c_t$=[R,G,B(Blue),C,M,Y,B(Black),W,N], $\sigma_t=-1$ for $t\in$ [1,2,3,4,5,6], and $\sigma_t=1$ for $t\in$ [7,8,9]. 
Note that $ROI(c_t)$ is the regions within this color.

The result is a color map that shows you saturation levels across the scene. Darker shades of gray are less saturated, and lighter shades are more saturated.

\subsection{Smoothing H}

As for $H_{smooth}$, we first convert RGB image into HSV space and set the V channel to 0.8, S channel to 0.5, then convert it back to RGB space.

\section{Intermediate Result Visualizations}
\label{intermediate}

\begin{figure*}[hthb]
\centering

\subcaptionbox{$M$}[0.13\linewidth]
{
\includegraphics[width=1\linewidth]{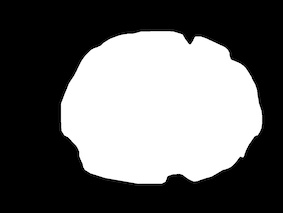}
\includegraphics[width=1\linewidth]{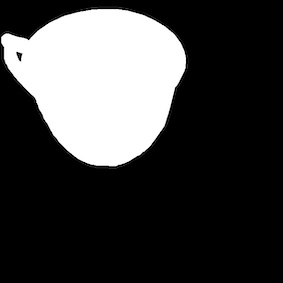}
\includegraphics[width=1\linewidth]{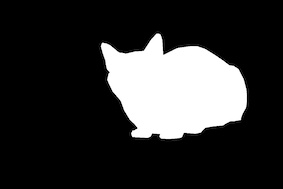}
\includegraphics[width=1\linewidth]{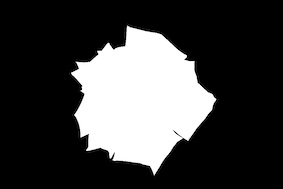}
\includegraphics[width=1\linewidth]{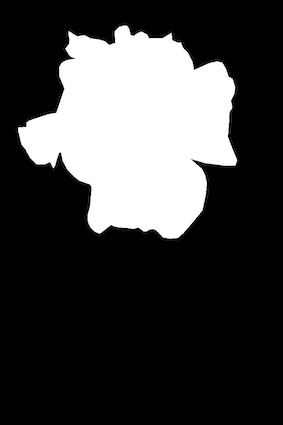}
\includegraphics[width=1\linewidth]{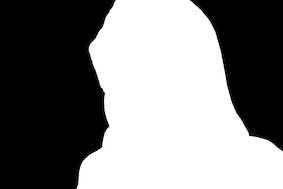}
}
\subcaptionbox{$I_c$}[0.13\linewidth]
{
\includegraphics[width=1\linewidth]{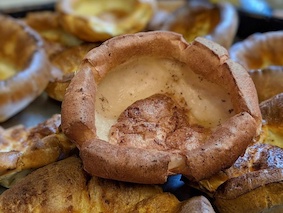}
\includegraphics[width=1\linewidth]{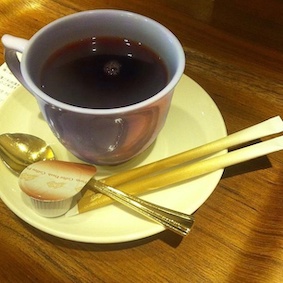}
\includegraphics[width=1\linewidth]{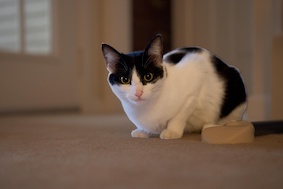}
\includegraphics[width=1\linewidth]{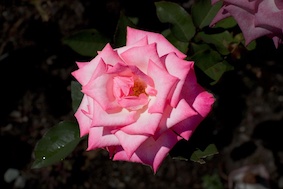}
\includegraphics[width=1\linewidth]{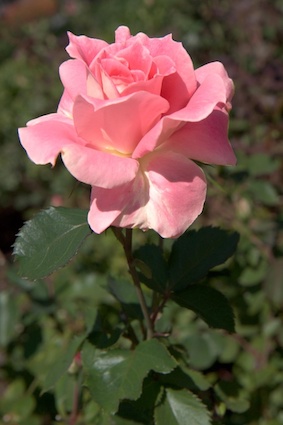}
\includegraphics[width=1\linewidth]{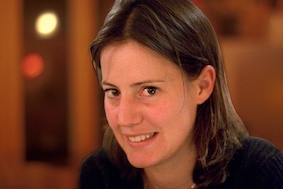}
}
\subcaptionbox{$I_1$}[0.13\linewidth]
{
\includegraphics[width=1\linewidth]{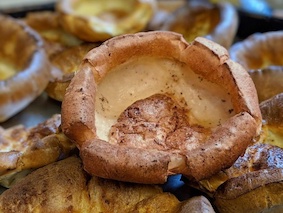}
\includegraphics[width=1\linewidth]{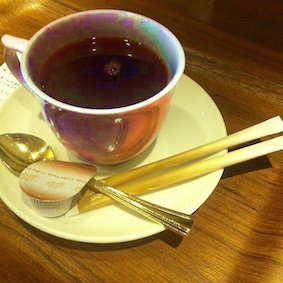}
\includegraphics[width=1\linewidth]{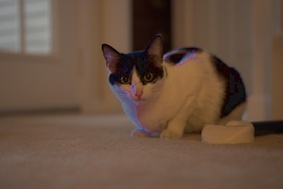}
\includegraphics[width=1\linewidth]{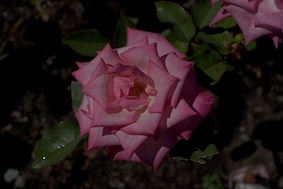}
\includegraphics[width=1\linewidth]{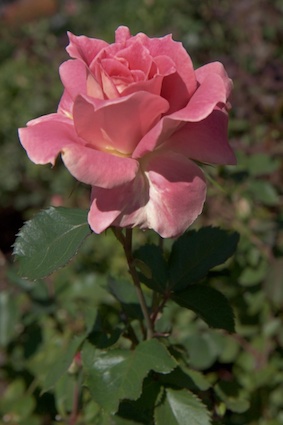}
\includegraphics[width=1\linewidth]{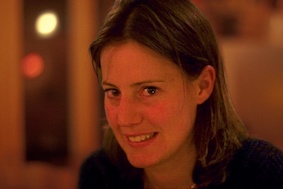}
}
\subcaptionbox{$I_2$}[0.13\linewidth]
{
\includegraphics[width=1\linewidth]{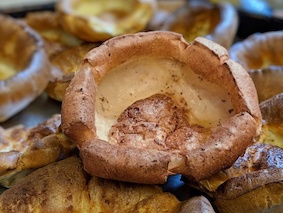}
\includegraphics[width=1\linewidth]{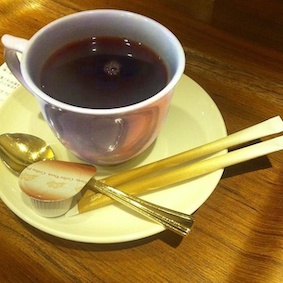}
\includegraphics[width=1\linewidth]{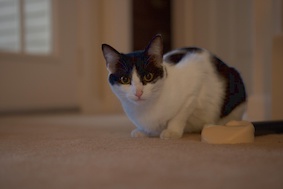}
\includegraphics[width=1\linewidth]{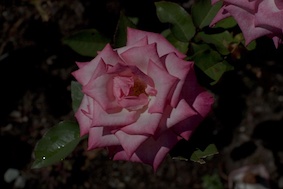}
\includegraphics[width=1\linewidth]{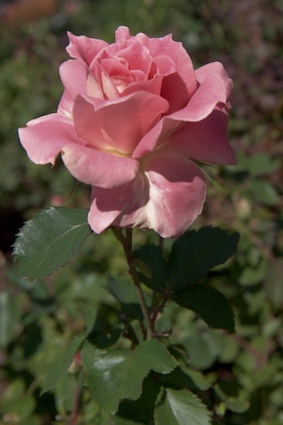}
\includegraphics[width=1\linewidth]{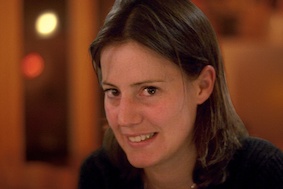}
}
\subcaptionbox{$I_3$}[0.13\linewidth]
{
\includegraphics[width=1\linewidth]{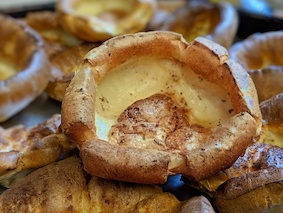}
\includegraphics[width=1\linewidth]{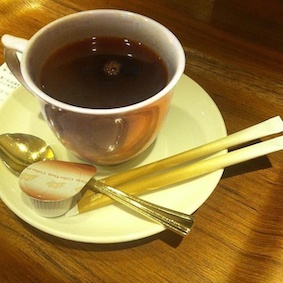}
\includegraphics[width=1\linewidth]{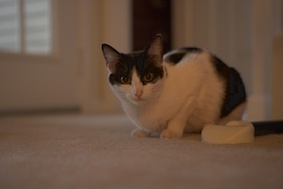}
\includegraphics[width=1\linewidth]{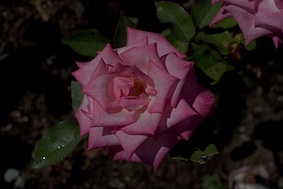}
\includegraphics[width=1\linewidth]{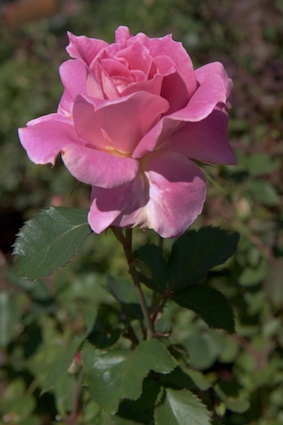}
\includegraphics[width=1\linewidth]{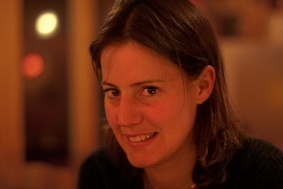}
}
\subcaptionbox{$I_4$}[0.13\linewidth]
{
\includegraphics[width=1\linewidth]{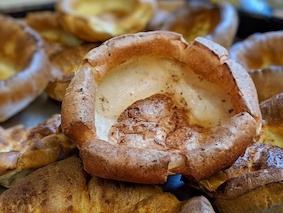}
\includegraphics[width=1\linewidth]{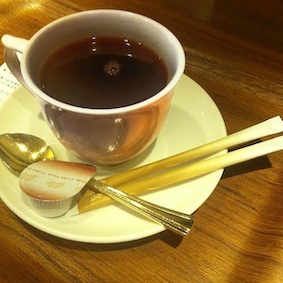}
\includegraphics[width=1\linewidth]{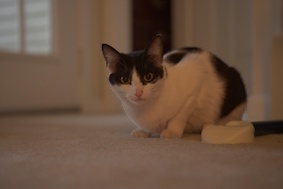}
\includegraphics[width=1\linewidth]{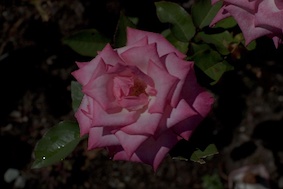}
\includegraphics[width=1\linewidth]{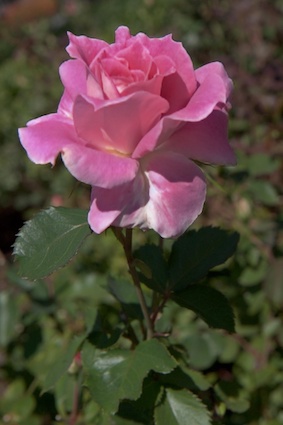}
\includegraphics[width=1\linewidth]{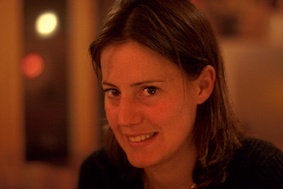}
}
\subcaptionbox{$I_{gt}$}[0.13\linewidth]
{
\includegraphics[width=1\linewidth]{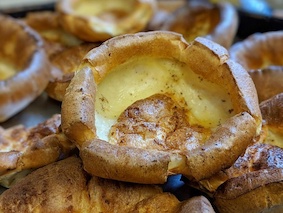}
\includegraphics[width=1\linewidth]{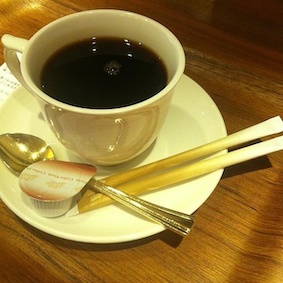}
\includegraphics[width=1\linewidth]{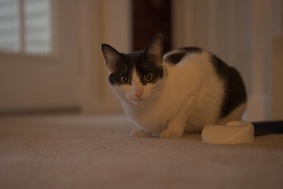}
\includegraphics[width=1\linewidth]{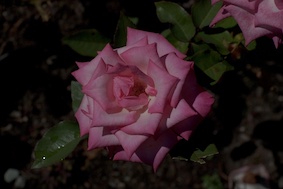}
\includegraphics[width=1\linewidth]{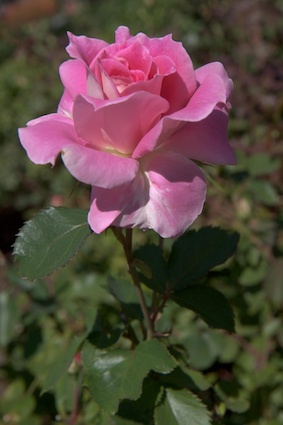}
\includegraphics[width=1\linewidth]{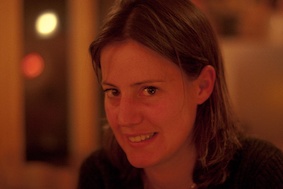}
}

\subcaptionbox{$V_c$}[0.1\linewidth]
{
\includegraphics[width=1\linewidth]{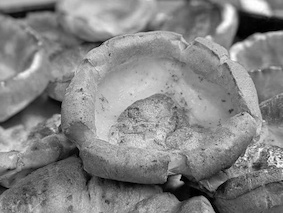}
\includegraphics[width=1\linewidth]{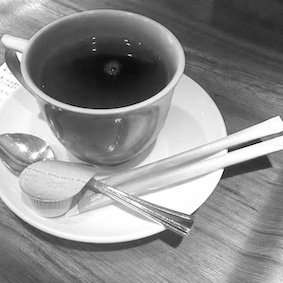}
\includegraphics[width=1\linewidth]{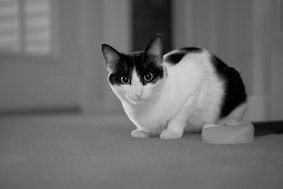}
\includegraphics[width=1\linewidth]{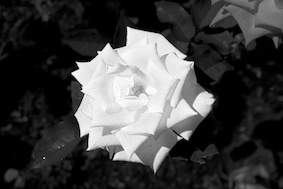}
\includegraphics[width=1\linewidth]{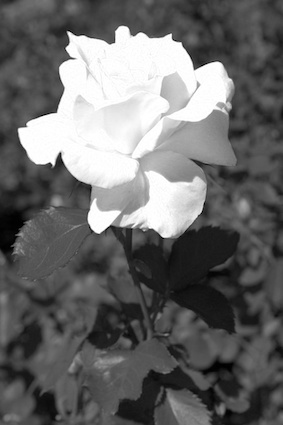}
\includegraphics[width=1\linewidth]{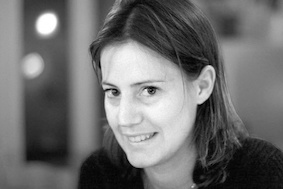}
}
\subcaptionbox{$V_1$}[0.1\linewidth]
{
\includegraphics[width=1\linewidth]{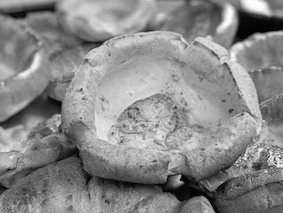}
\includegraphics[width=1\linewidth]{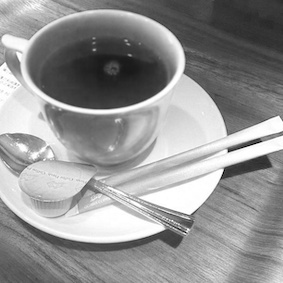}
\includegraphics[width=1\linewidth]{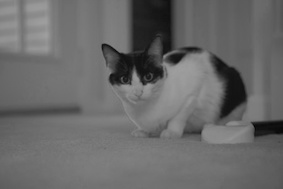}
\includegraphics[width=1\linewidth]{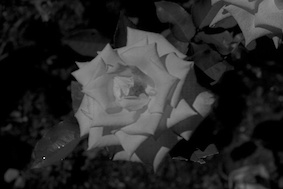}
\includegraphics[width=1\linewidth]{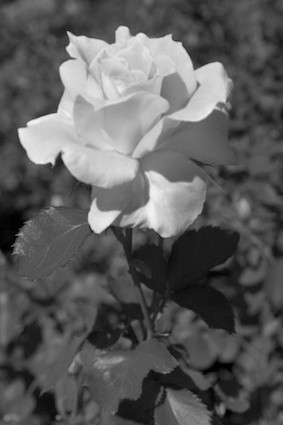}
\includegraphics[width=1\linewidth]{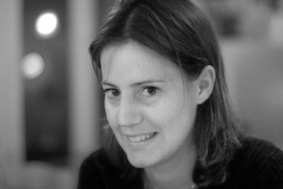}
}
\subcaptionbox{$V_{gt}$}[0.1\linewidth]
{
\includegraphics[width=1\linewidth]{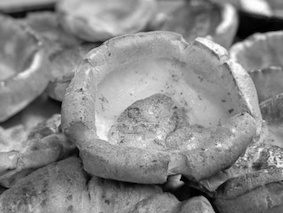}
\includegraphics[width=1\linewidth]{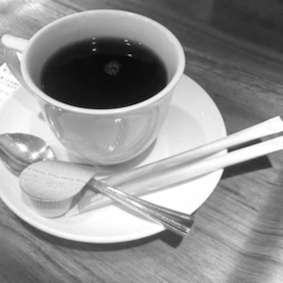}
\includegraphics[width=1\linewidth]{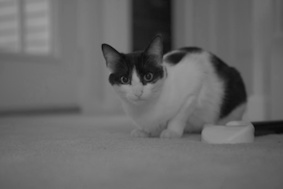}
\includegraphics[width=1\linewidth]{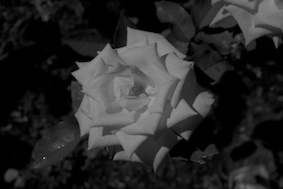}
\includegraphics[width=1\linewidth]{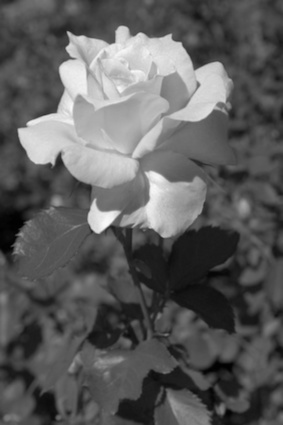}
\includegraphics[width=1\linewidth]{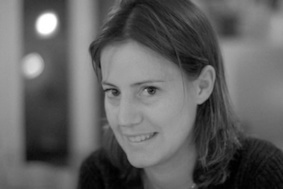}
}
\subcaptionbox{$S_c$}[0.1\linewidth]
{
\includegraphics[width=1\linewidth]{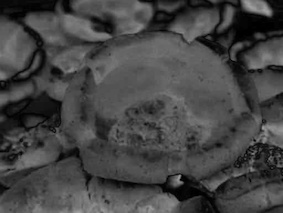}
\includegraphics[width=1\linewidth]{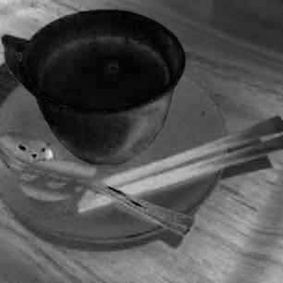}
\includegraphics[width=1\linewidth]{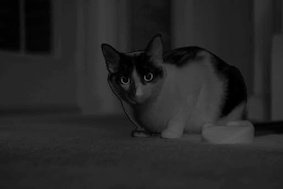}
\includegraphics[width=1\linewidth]{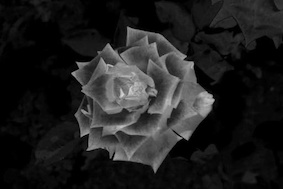}
\includegraphics[width=1\linewidth]{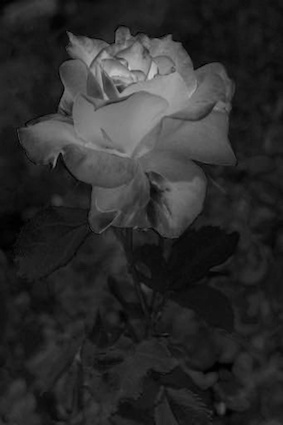}
\includegraphics[width=1\linewidth]{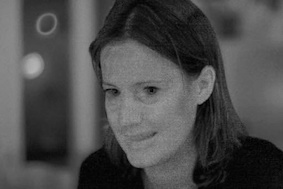}
}
\subcaptionbox{$S_1$}[0.1\linewidth]
{
\includegraphics[width=1\linewidth]{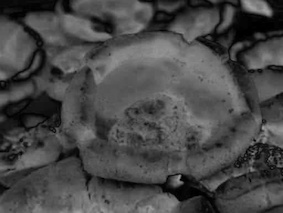}
\includegraphics[width=1\linewidth]{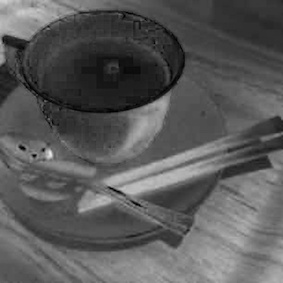}
\includegraphics[width=1\linewidth]{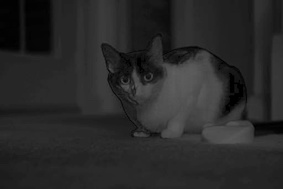}
\includegraphics[width=1\linewidth]{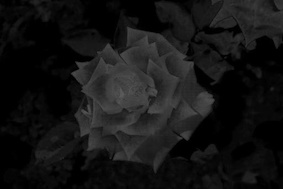}
\includegraphics[width=1\linewidth]{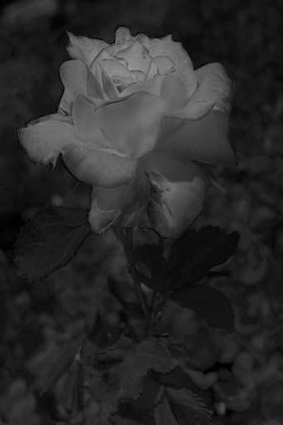}
\includegraphics[width=1\linewidth]{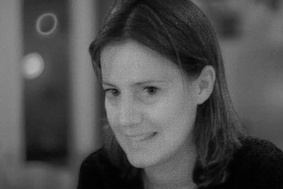}
}
\subcaptionbox{$S_{gt}$}[0.1\linewidth]
{
\includegraphics[width=1\linewidth]{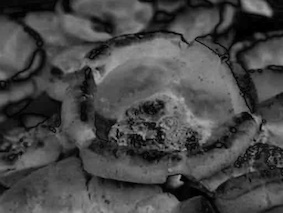}
\includegraphics[width=1\linewidth]{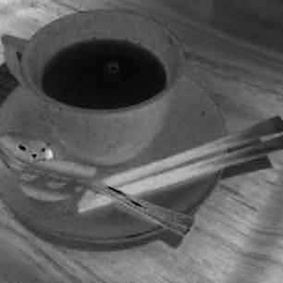}
\includegraphics[width=1\linewidth]{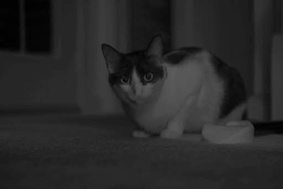}
\includegraphics[width=1\linewidth]{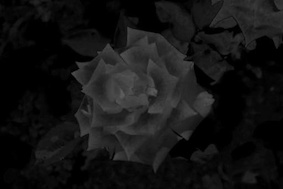}
\includegraphics[width=1\linewidth]{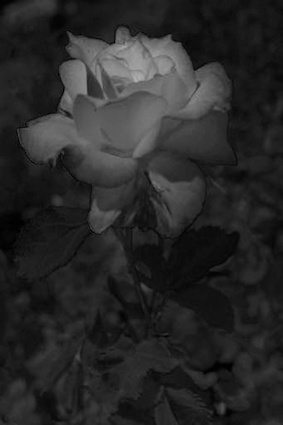}
\includegraphics[width=1\linewidth]{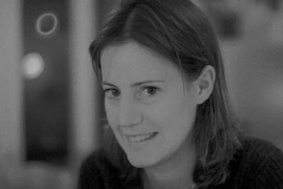}
}
\subcaptionbox{$H_c$}[0.1\linewidth]
{
\includegraphics[width=1\linewidth]{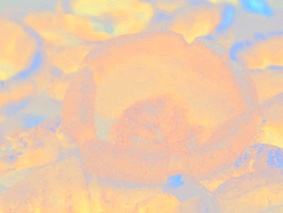}
\includegraphics[width=1\linewidth]{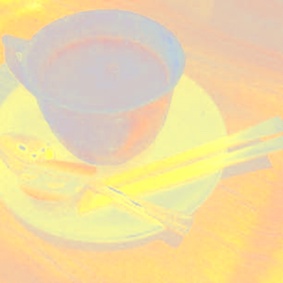}
\includegraphics[width=1\linewidth]{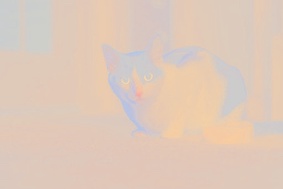}
\includegraphics[width=1\linewidth]{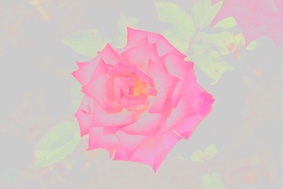}
\includegraphics[width=1\linewidth]{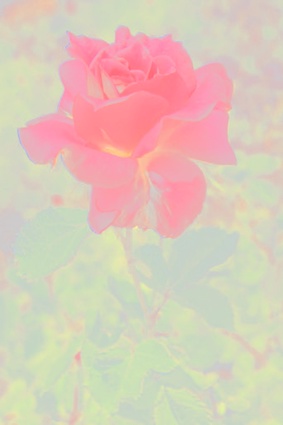}
\includegraphics[width=1\linewidth]{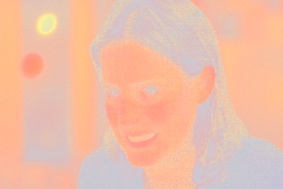}
}
\subcaptionbox{$H_1$}[0.1\linewidth]
{
\includegraphics[width=1\linewidth]{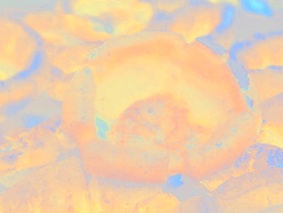}
\includegraphics[width=1\linewidth]{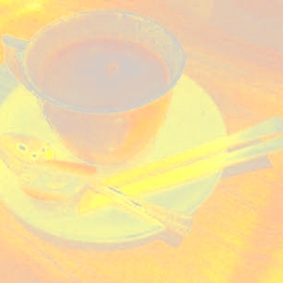}
\includegraphics[width=1\linewidth]{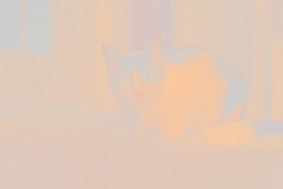}
\includegraphics[width=1\linewidth]{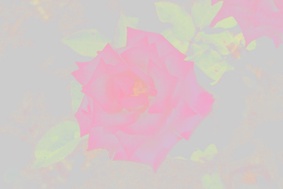}
\includegraphics[width=1\linewidth]{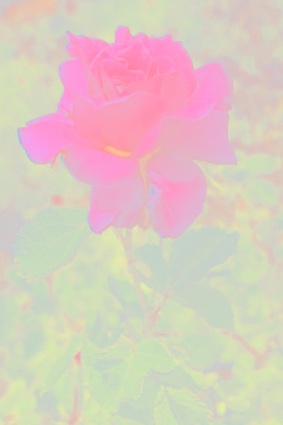}
\includegraphics[width=1\linewidth]{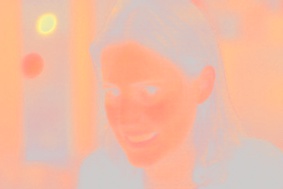}
}
\subcaptionbox{$H_{gt}$}[0.1\linewidth]
{
\includegraphics[width=1\linewidth]{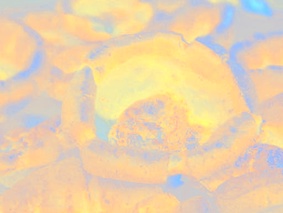}
\includegraphics[width=1\linewidth]{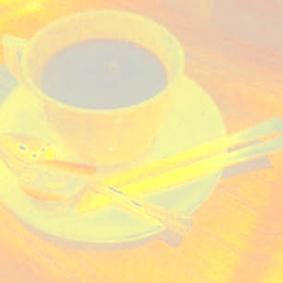}
\includegraphics[width=1\linewidth]{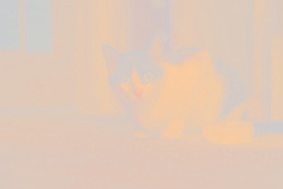}
\includegraphics[width=1\linewidth]{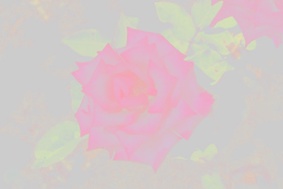}
\includegraphics[width=1\linewidth]{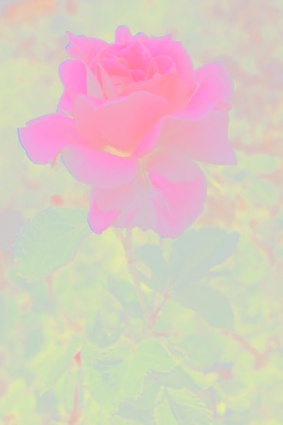}
\includegraphics[width=1\linewidth]{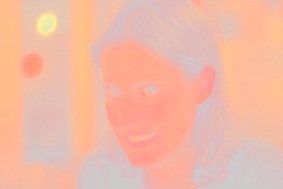}
}
\caption{Intermediate results.}
\label{fig:im_result}
\end{figure*}

We show that our intermediate result of DICF matches their design purpose in Fig.~\ref{fig:im_result}. 
$I_c$ is the input image. $I_1,I_2,I_3,I_4$ are outputs after $F_{val},F_{sat},F_{hue},F_{attn}$. 
$H_c,S_c,V_c$ are input's HSV maps. 
$H_3, S_2, V_1$ are their harmonized counterparts.
$I_{gt},V_{gt},S_{gt}, H_{gt}$ are ground truths.

As illustrated in the upper part of Fig.~\ref{fig:im_result}, each intermediate result of $I_1=F_{val}(I_c)$, $I_2=F_{sat}(I_1)$, $I_3=F_{hue}(I_2)$ not only changes its corresponding image attributes: \textit{value}, \textit{saturation} and \textit{hue}, but also maintains reasonable visual quality.
This is untrivial since we didn't apply direct RGB loss on $I_1$,$I_2$,$I_3$, instead we only apply auxiliary HSV loss on its specific channels.
This means that users can choose any of $I_1, I_2, I_3, I_4$ as their desired output if they only want to change part of these attributes, while previous works only provides a final result which is $I_4$ in our framework.
This brings more flexility and robustness for users when approaching DCCF.

We further illustrate the comprehensibility in the bottom part of Fig.~\ref{fig:im_result}, the modified channel is closer to ground truth after the operation of DCCF.
This also proves the effect of HSV loss in our framework.

\section{Comprehensible Interaction}
\label{interaction}

\begin{figure}[hthb]
\centering

\begin{subfigure}[t]{1\linewidth}\label{fig:edit_filter_L}
\includegraphics[width=1\textwidth]{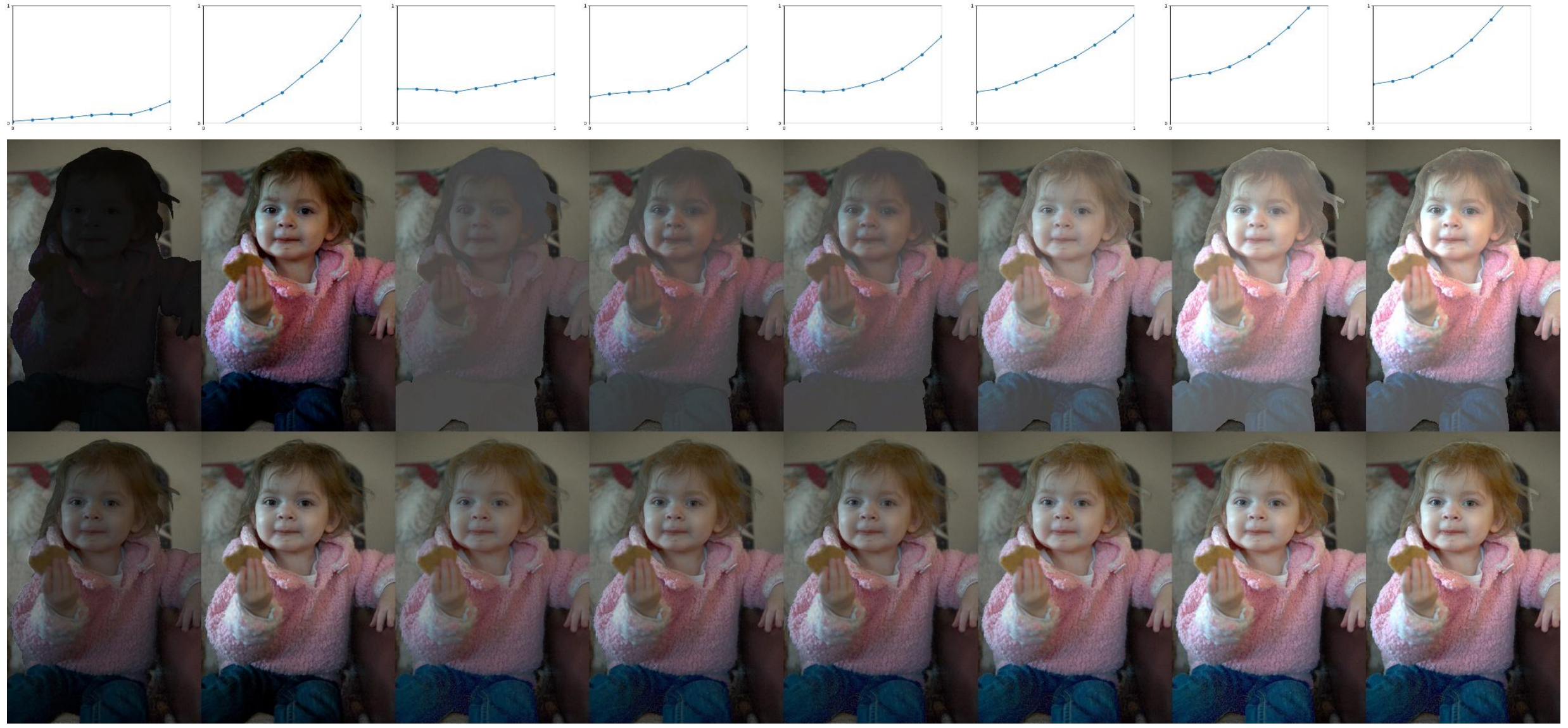}
\caption{Illustration of interactive value adjustment. }
\end{subfigure}

\begin{subfigure}[t]{1\linewidth}\label{fig:edit_filter_S}
\includegraphics[width=1\textwidth]{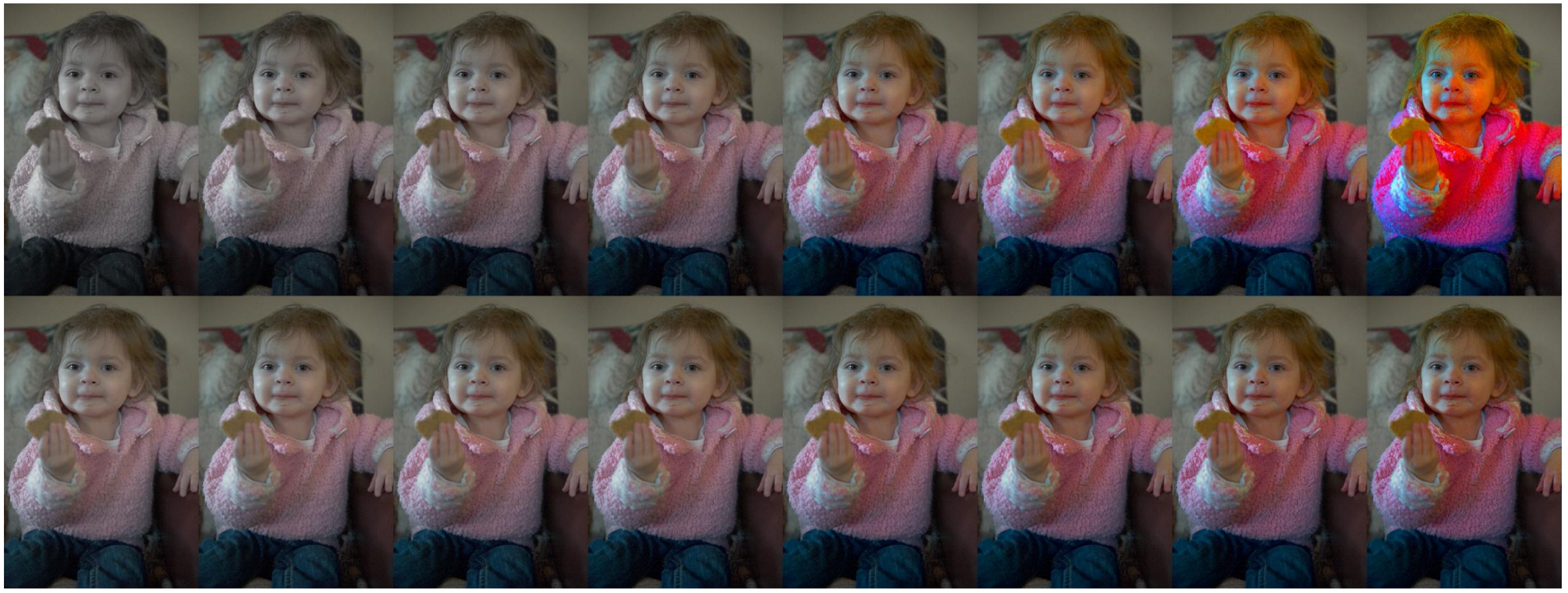}
\caption{Illustration of interactive saturation adjustment. }
\end{subfigure}

\caption{Interactive adjustment.
Upper: users' global adjustment. Bottom: interactive adjustment with DCCF.
}
\label{fig:edit_filter_LS}
\vspace{-10pt}
\end{figure}

To interact with $F_{val}$, in standard image processing softwares, users need to provide a curve to adjust value.
In our framework, users can set $[\phi_0,...\phi_m]$ to approximate a curve.
As shown in Fig.~\ref{fig:edit_filter_LS}(a), the first row is the visualization of several curves provided by user.
The second row is the result of directly applying these curves on RGB images which degrades image quality since each pixel has the same tuning curve.
The third row is the result of applying a weighted fusion($\alpha=0.5$) of user's global curve and DCCF's filter map $F_{val}$.

To interact with $F_{sat}$, users can change image saturation by $\sigma\in[-1, 1]$. 
As shown in Fig.~\ref{fig:edit_filter_LS}(b), we set a series of $\sigma\in[-0.8,-0.6,-0.4,-0.2,0.2,0.4,0.6,0.8]$ to generate the first row, which is a global adjustment and each pixel has the same $\sigma$ and could suffer from over-saturation.
The second row is the weighted fusion($\alpha=0.5$) of user's $\sigma$ and DCCF's filter map $F_{sat}$.

\section{More High-Resolution Visualizations}
\label{sec:high-res-vis}
\begin{figure*}[hthb]
\centering
\subcaptionbox{Mask}[0.13\linewidth]
{
\includegraphics[width=1\linewidth]{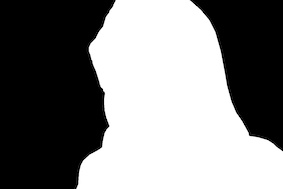}
\includegraphics[width=1\linewidth]{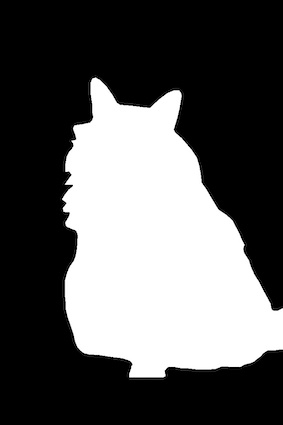}
\includegraphics[width=1\linewidth]{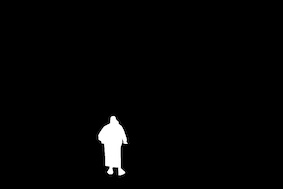}
\includegraphics[width=1\linewidth]{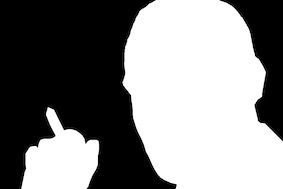}
\includegraphics[width=1\linewidth]{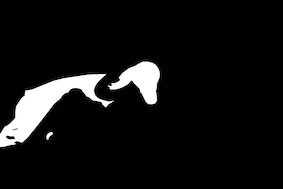}
\includegraphics[width=1\linewidth]{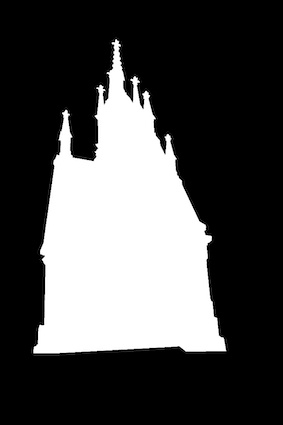}
\includegraphics[width=1\linewidth]{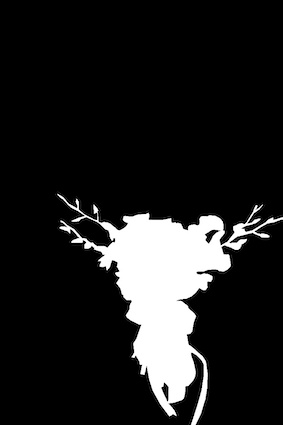}
\includegraphics[width=1\linewidth]{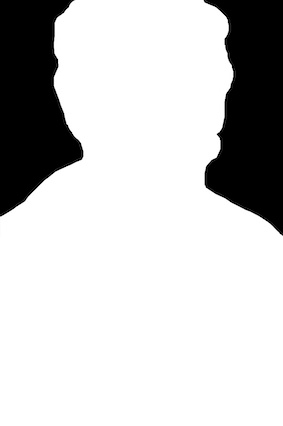}
}
\subcaptionbox{Input}[0.13\linewidth]
{
\includegraphics[width=1\linewidth]{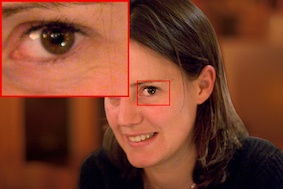}
\includegraphics[width=1\linewidth]{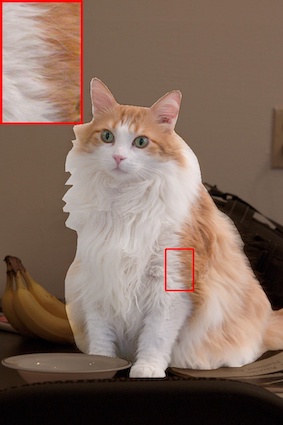}
\includegraphics[width=1\linewidth]{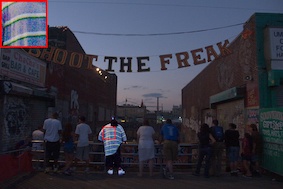}
\includegraphics[width=1\linewidth]{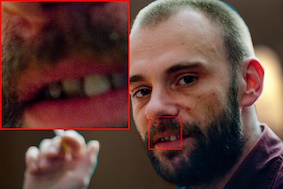}
\includegraphics[width=1\linewidth]{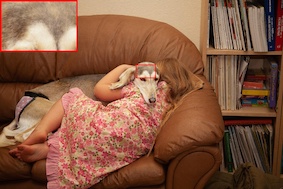}
\includegraphics[width=1\linewidth]{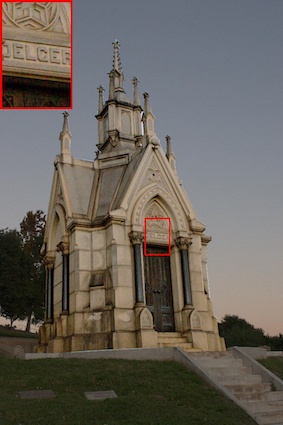}
\includegraphics[width=1\linewidth]{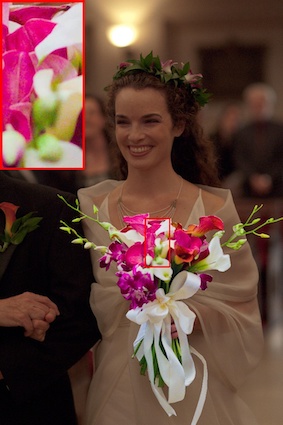}
\includegraphics[width=1\linewidth]{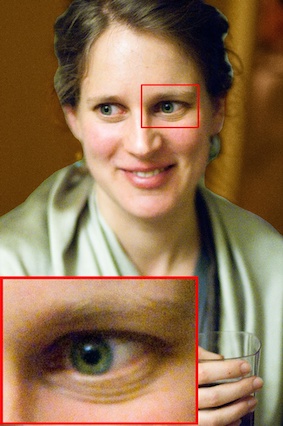}
}
\subcaptionbox{BU}[0.13\linewidth]
{
\includegraphics[width=1\linewidth]{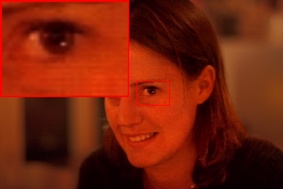}
\includegraphics[width=1\linewidth]{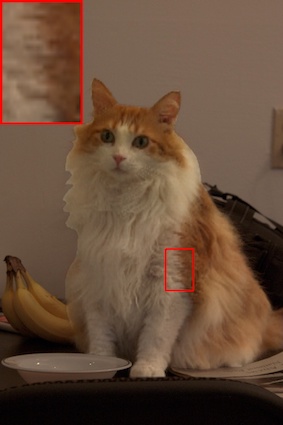}
\includegraphics[width=1\linewidth]{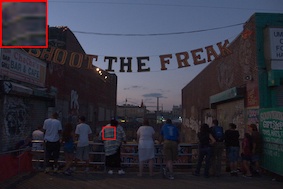}
\includegraphics[width=1\linewidth]{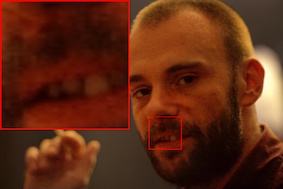}
\includegraphics[width=1\linewidth]{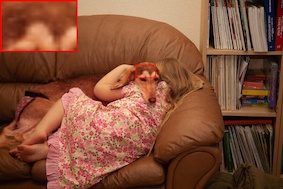}
\includegraphics[width=1\linewidth]{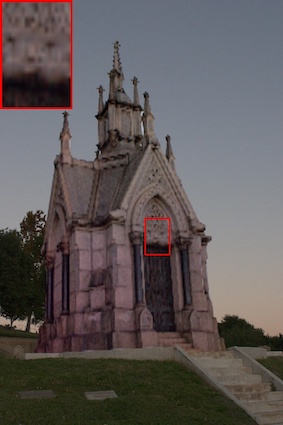}
\includegraphics[width=1\linewidth]{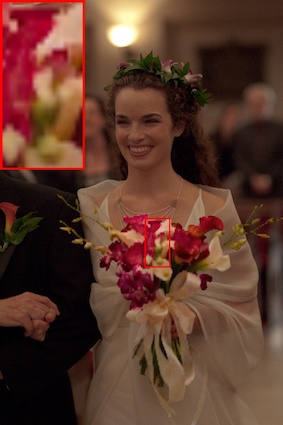}
\includegraphics[width=1\linewidth]{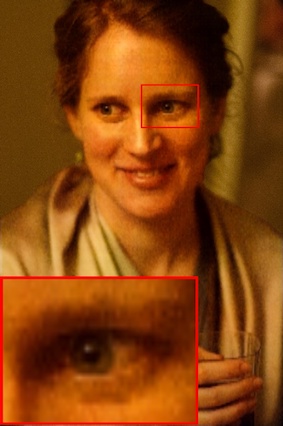}
}
\subcaptionbox{GF\cite{hekaiming2013GF}}[0.13\linewidth]
{
\includegraphics[width=1\linewidth]{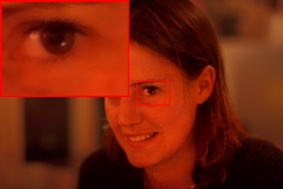}
\includegraphics[width=1\linewidth]{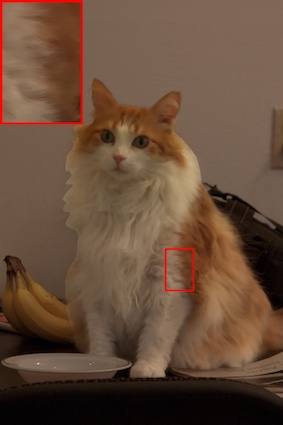}
\includegraphics[width=1\linewidth]{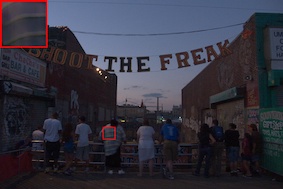}
\includegraphics[width=1\linewidth]{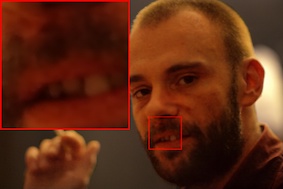}
\includegraphics[width=1\linewidth]{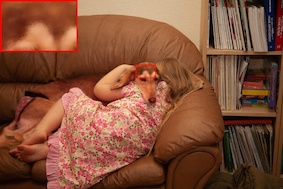}
\includegraphics[width=1\linewidth]{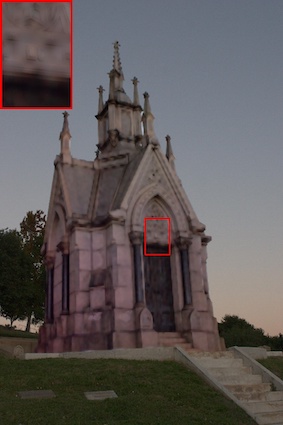}
\includegraphics[width=1\linewidth]{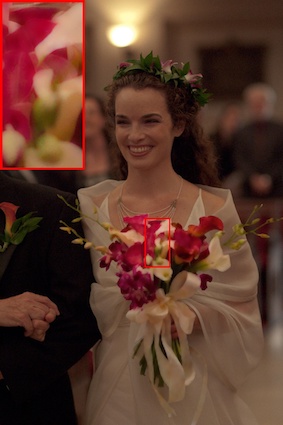}
\includegraphics[width=1\linewidth]{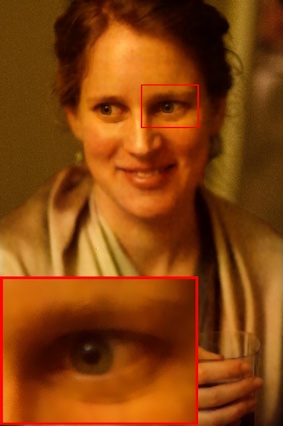}
}
\subcaptionbox{BGU\cite{chen2016bilateral}}[0.13\linewidth]
{
\includegraphics[width=1\linewidth]{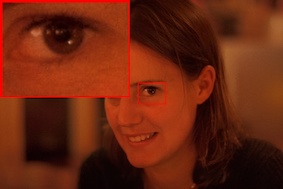}
\includegraphics[width=1\linewidth]{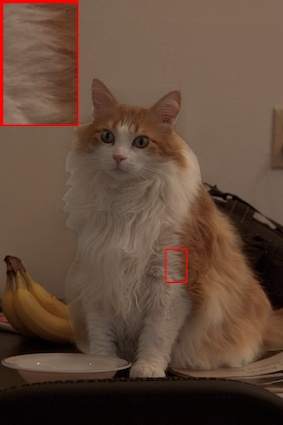}
\includegraphics[width=1\linewidth]{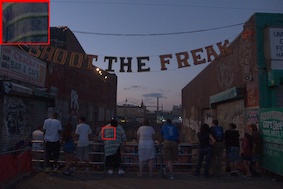}
\includegraphics[width=1\linewidth]{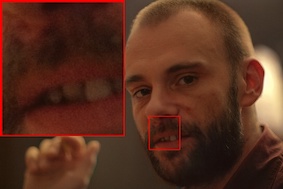}
\includegraphics[width=1\linewidth]{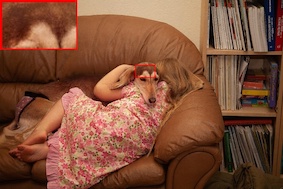}
\includegraphics[width=1\linewidth]{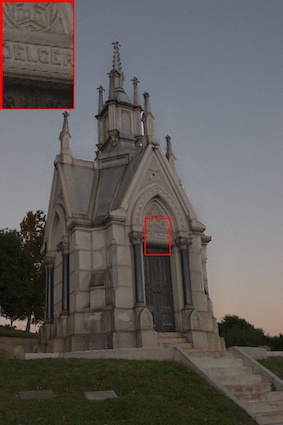}
\includegraphics[width=1\linewidth]{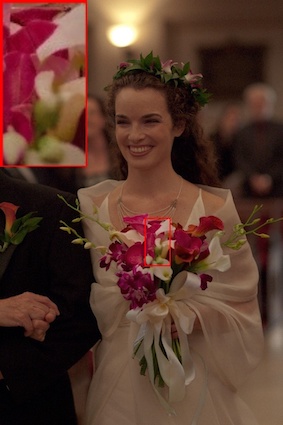}
\includegraphics[width=1\linewidth]{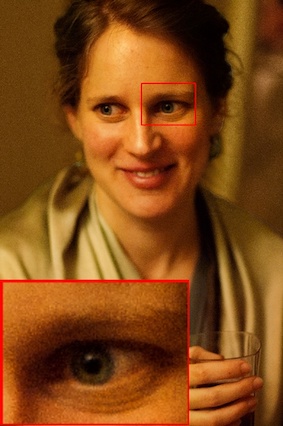}
}
\subcaptionbox{Ours}[0.13\linewidth]
{
\includegraphics[width=1\linewidth]{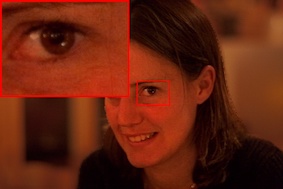}
\includegraphics[width=1\linewidth]{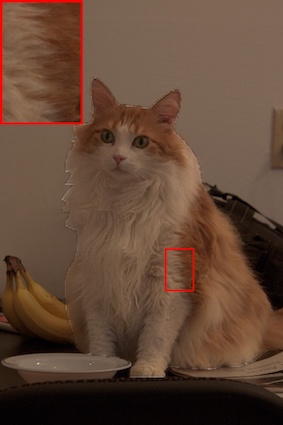}
\includegraphics[width=1\linewidth]{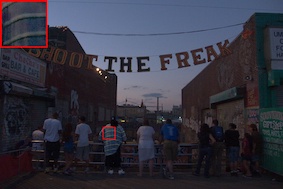}
\includegraphics[width=1\linewidth]{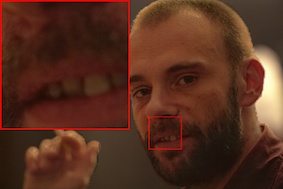}
\includegraphics[width=1\linewidth]{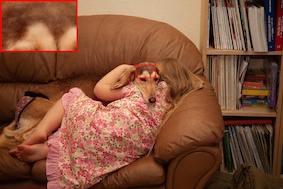}
\includegraphics[width=1\linewidth]{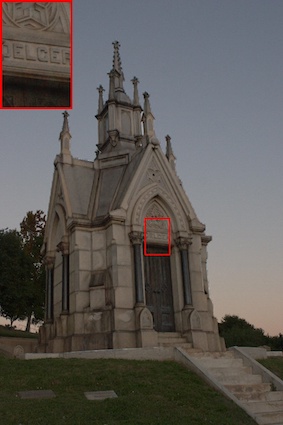}
\includegraphics[width=1\linewidth]{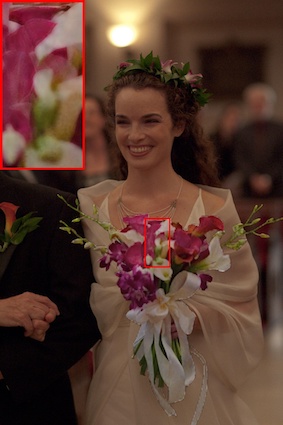}
\includegraphics[width=1\linewidth]{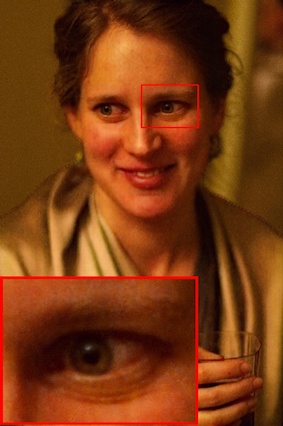}
}
\subcaptionbox{GT}[0.13\linewidth]
{
\includegraphics[width=1\linewidth]{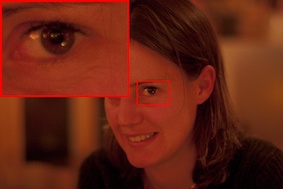}
\includegraphics[width=1\linewidth]{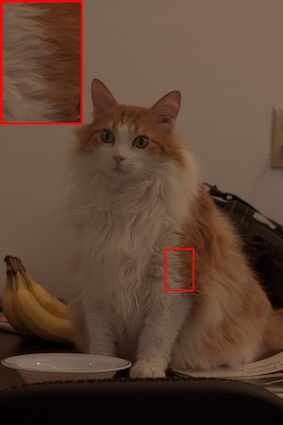}
\includegraphics[width=1\linewidth]{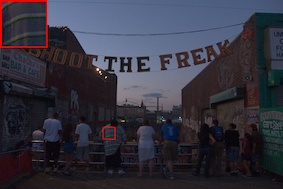}
\includegraphics[width=1\linewidth]{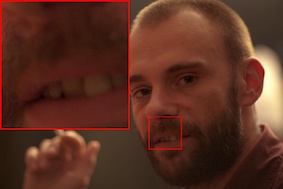}
\includegraphics[width=1\linewidth]{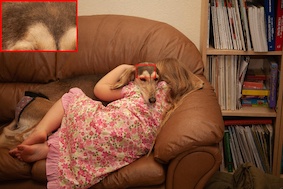}
\includegraphics[width=1\linewidth]{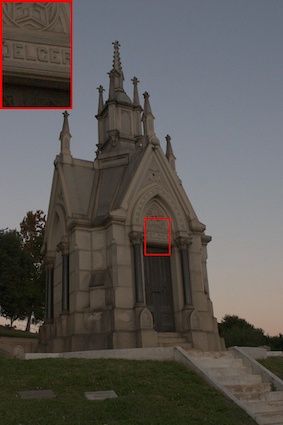}
\includegraphics[width=1\linewidth]{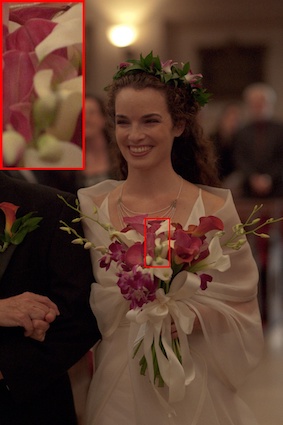}
\includegraphics[width=1\linewidth]{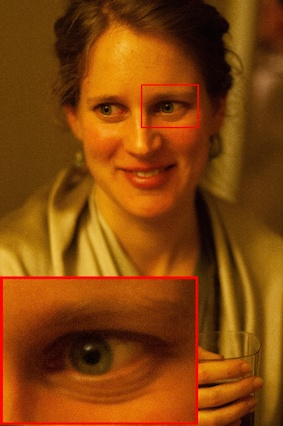}
}
\caption{Visualization of high-resolution results. }

\label{fig:HR_vis_supp}
\end{figure*}

We provide more visualizations of final results compared with previous methods in Fig.~\ref{fig:HR_vis_supp}.
Since DCCF is an end-to-end framework, it has strong transformation capability while maintaining high-resolution details.

\section{Limitations}
\label{sec:limitation}

\begin{figure}[htbp]
  \centering
  
  \begin{subfigure}[t]{0.3\linewidth}
  \centering
  \includegraphics[width=1\linewidth]{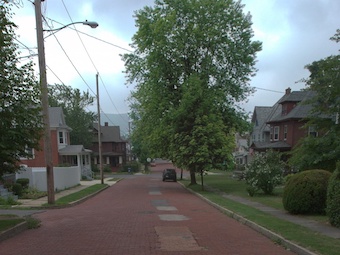}
  \end{subfigure}
  \begin{subfigure}[t]{0.3\linewidth}
  \centering
  \includegraphics[width=1\linewidth]{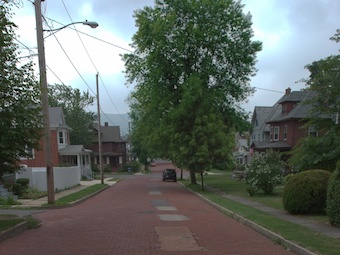}
  \end{subfigure}
  \begin{subfigure}[t]{0.3\linewidth}
  \centering
  \includegraphics[width=1\linewidth]{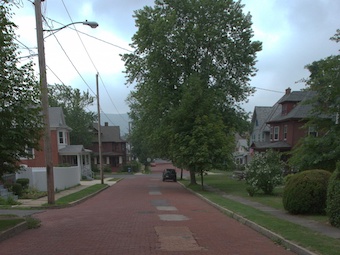}
  \end{subfigure}

  \begin{subfigure}[t]{0.3\linewidth}
    \centering
    \includegraphics[width=1\linewidth]{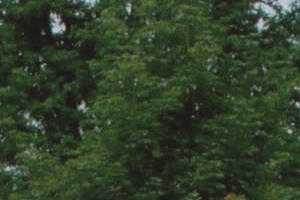}
    \caption{BGU~\cite{chen2016bilateral}}
    \end{subfigure}
    \begin{subfigure}[t]{0.3\linewidth}
    \centering
    \includegraphics[width=1\linewidth]{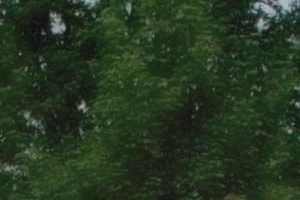}
    \caption{DCCF}
    \end{subfigure}
    \begin{subfigure}[t]{0.3\linewidth}
    \centering
    \includegraphics[width=1\linewidth]{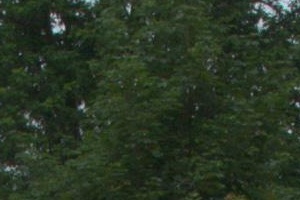}
    \caption{GT}
    \end{subfigure}
  
  \caption{Potential limitation. The second row is amplified details. It is observed that even DCCF learns better color adjustment, the detail starts to blur on high frequency regions like \textit{leaf}.}
  
  \label{fig:limitation}
  \end{figure}

Since the high resolution result is guided by low resolution stream in our framework, the claim of insensitivity to resolution is valid only if the processed image has enough information shared across all signal frequency. 
In the case of extremely high frequency contents, it may fail to reharmonize images properly, see Fig~\ref{fig:limitation}.

\newcommand{\imdir}{figures/intermediate_result_supp/}
\newcommand{\ima}{f2598_1_2}
\newcommand{\imb}{f1570_1_1}
\newcommand{\imc}{a3163_1_1}
\newcommand{\imd}{a0041_1_5}
\newcommand{\ime}{a0782_1_4}
\newcommand{\imf}{a3427_1_4}

\newcommand{\hrdir}{figures/hr_details}
\newcommand{\hra}{a3427_1_4}
\newcommand{\hrb}{a0196_1_4}
\newcommand{\hrc}{a4732_1_1}
\newcommand{\hrd}{a3977_1_5}
\newcommand{\hre}{a4696_1_5}
\newcommand{\hrf}{a0249_1_5}
\newcommand{\hrg}{a0779_1_3}
\newcommand{\hrh}{a1580_1_4}

\end{document}